\pdfoutput=1

\documentclass{article}

\usepackage{xcolor}
\usepackage{makecell}
\usepackage{booktabs,array,footnote}
\usepackage{hyperref}
\usepackage[nonatbib]{neurips_2023}

\usepackage{times}
\usepackage{latexsym}
\usepackage{graphicx}
\usepackage{amsmath}
\usepackage{amssymb}
\usepackage{booktabs}
\usepackage{multirow}
\usepackage{enumitem}
\newcommand{\red}[1]{\textcolor{red}{#1}}

\usepackage{xspace}
\usepackage[T1]{fontenc}
\usepackage{tabularx}

\usepackage{wrapfig}
\usepackage{tabulary}

\usepackage[utf8]{inputenc}
\newcommand{\nop}[1]{}
\usepackage{listings}
\usepackage{caption}
\usepackage{subcaption}

\usepackage{microtype}

\usepackage{inconsolata}
\usepackage{longtable}
\usepackage{caption}
\usepackage{wrapfig}
\usepackage{sidecap}
\title{PaLI-X: On Scaling up a Multilingual Vision and Language Model}

\begin{document}

\maketitle
\newcommand{\NAME}{PaLI\xspace}
\newcommand{\NEWNAME}{PaLI-X\xspace}
\newcommand{\NEWSIZE}{55B}
\newcommand{\PTDATA}{WebLI\xspace}

\newcommand{\prompt}[1]{"\emph{#1}"}
\newcommand{\exid}{$\langle\mathrm{extra\_id\_0}\rangle$\xspace}
\newcommand{\exidk}{$\langle\mathrm{extra\_id}\_k\rangle$\xspace}

\newcommand{\lang}{$\langle\mathrm{lang}\rangle$\xspace}
\newcommand{\pos}{$\langle\mathrm{pos}\rangle$\xspace}
\newcommand{\capfirst}{$\langle\mathrm{cap_1}\rangle$\xspace}
\newcommand{\capsecond}{$\langle\mathrm{cap_2}\rangle$\xspace}
\newcommand{\ocrtext}{$\langle\mathrm{OCR\_text}\rangle$\xspace}
\newcommand{\obj}{$\langle\mathrm{object_1}\rangle$\xspace}
\newcommand{\objk}{$\langle\mathrm{object_k}\rangle$\xspace}
\newcommand{\objn}{$\langle\mathrm{object_N}\rangle$\xspace}

\newcommand{\vqaset}{VQA2.0\xspace}
\newcommand{\gqa}{GQA\xspace}
\newcommand{\okvqa}{OKVQA\xspace}
\newcommand{\vsw}{Visual7W\xspace}
\newcommand{\cocoqa}{COCOQA\xspace}
\newcommand{\advqa}{AdVQA\xspace}
\newcommand{\coco}{COCO\xspace}
\newcommand{\cococite}{\cite{lin14eccv,chen15arxiv}}
\newcommand{\oid}{Open~Images}
\newcommand{\oidcite}{\cite{kuznetsova20ijcv,openimages}}
\newcommand{\flickr}{Flickr30k}
\newcommand{\flickrcite}{\cite{young14tacl,plummer17ijcv}}
\newcommand{\ade}{ADE20K}
\newcommand{\cc}{Conceptual Captions}
\newcommand{\ccshort}{CC}
\newcommand{\ccslong}{CC3M\xspace}
\newcommand{\cccite}{\cite{cc3m}}

\newcommand{\qsq}{$\mathrm{VQ}^{2}\!\mathrm{A}$\xspace}
\newcommand{\qsqcc}{\qsq-CC3M\xspace}
\newcommand{\qsqcoco}{\qsq-COCO\xspace}
\newcommand{\maxm}{$\mathrm{MaXM}$\xspace}
\newcommand{\mvr}{$\mathrm{MAVERICS}$\xspace}
\newcommand{\xmodal}{$\mathrm{XM3600}$\xspace}
\newcommand{\XM}{$\mathrm{XM3600}$\xspace}

\newcommand{\imres}[1]{#1$\times$#1}

\newcommand{\sotamodel}[1]{\textcolor{gray}{{\scriptsize #1}}}

\begin{abstract}
We present the training recipe and results of scaling up \NEWNAME, a multilingual vision and language model,
both in terms of size of the components and the breadth of its training task mixture.
Our model achieves new levels of performance on a wide-range of varied and complex tasks, including multiple image-based captioning and question-answering tasks, image-based document understanding and few-shot (in-context) learning, as well as object detection, video question answering, and video captioning. 
\NEWNAME advances the state-of-the-art on most vision-and-language  benchmarks considered (25+ of them). 
Finally, we observe emerging capabilities, such as complex counting and multilingual object detection, tasks that are not explicitly in the training mix.
\end{abstract}
\section{Introduction}
The success of scaling language models~\cite{chowdhery2022palm, gpt3, glam, anil2023palm} makes it appealing to similarly scale Vision-Language (V\&L) models, and investigate the improvements,  capabilities, and  emergent properties of such models.
Inspired by the work in~\cite{pali2}, we present \NEWNAME, a multilingual vision and language model with reusable scaled-up components, consisting of a pretrained large-capacity visual encoder (using~\cite{vit-22b} as the starting point) and  a pretrained language-only encoder-decoder (using~\cite{tay2023ul2} as the starting point), further trained at-scale on a vision-and-language data mixture using a combination of self-supervision and full-supervision signals.

\begin{figure}
\centering
\includegraphics[width=\textwidth]{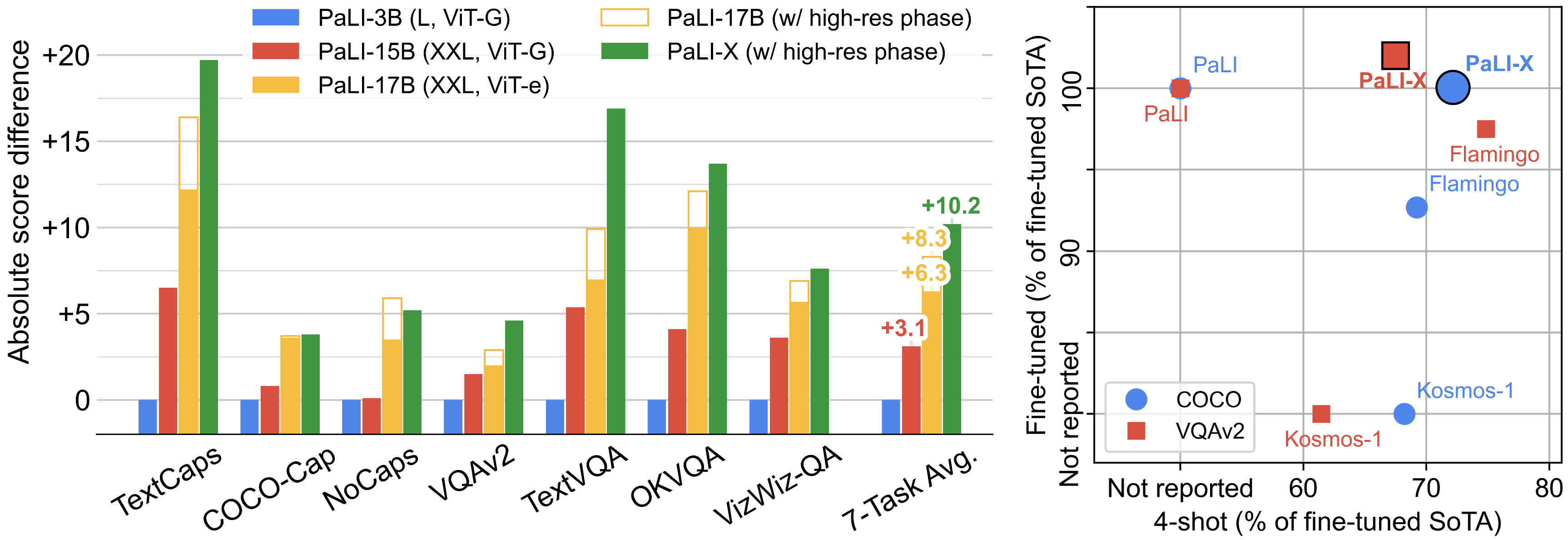}
\caption{[Left] Comparing \NEWNAME against \NAME on image-captioning and VQA benchmarks. [Right] The Pareto frontier between few-shot and fine-tuned performance, comparing \NEWNAME with PaLI~\cite{pali2}, Flamingo~\cite{alayrac2022flamingo}, and Kosmos-1~\cite{huang2023language}.
}
\label{fig:scaling-main}
\end{figure}

One clear pattern that emerges from the combination of results from \NAME~\cite{pali2} and the work we present in this paper is that scaling \emph{both} V\&L components together brings increases in performance across a wide range of tasks.
We show this by comparing against the same benchmarks used for \NAME (Fig. \ref{fig:scaling-main}, Left), and also against new benchmarks for which the new capabilities of \NEWNAME are evaluated (e.g., ChartQA, AI2D, DocVQA, InfographicVQA, as well as video understanding tasks).
We observe that scaling leads to large improvements over the results of the \NAME model, and also over specialized large-scale models that are trained specifically to solve certain tasks, often with the help of (often much larger) text-only LLMs~\cite{deplot}.
In particular, we find that increasing both the effective capacity of the vision component (which~\cite{wang2022git} does more unilaterally) and of the language component (which~\cite{alayrac2022flamingo} also does unilaterally) is beneficial; the new \NEWNAME model provides more balanced parameter allocation than any other prior work (roughly 40\%-60\% split of the total capacity).

Aside from confirming the impact of scale, the original contribution of \NEWNAME consists in leveraging the mixture-of-objectives proposed in~\cite{tay2023ul2} for vision-and-language modeling, and showing that it results in a model that improves both state-of-the-art results and the Pareto frontier for fine-tuning and few-shot configurations (Fig. \ref{fig:scaling-main}, Right).

We also observe emergent properties based on \NEWNAME's results compared to previous models with similar architecture but smaller sizes.
For instance, we report drastically improved performance on the counting ability (See Table~\ref{table:captioning} and Appendix~\ref{appendix:image_results}), both for the plain variety (count all instances of a class) and the complex variety (count instances based on a natural language description), that are not attributable to training design\footnote{Plain counting is usually achievable via good object detection, while complex counting requires a fine-grained understanding of the alignment between language-based specifications and visually-based occurrences.}.
Additionally, we present qualitative insights into the model's performance (Appendix~\ref{appendix:model_info_and_example}), with an emphasis on multilingual transfer learning such as the ability to detect objects using non-English labels (Fig. \ref{fig:det_examples}), and the ability to switch between the language of text present in the image (e.g., English) and the language of the generated image caption (e.g., Romanian).

 Our technical contributions include the following:
 \begin{enumerate}
     \item 
     We scale a Vision-Language model to achieve outstanding performance on a wide variety of benchmarks. We observe that scaling {\it both} the Vision \& Language components is advantageous and report that performance remains unsaturated at this scale.     
     \item We show that training such a model with a mixture of objectives that combines prefix-completion and masked-token completion improves the Pareto frontier for fine-tuning vs few-shot performance at this scale.
     \item We show that a high-capacity vision encoder (ViT-22B) can be effectively co-trained for image classification and OCR label classification\footnote{We use OCR tokens produced by the GCP Vision API over the training images as targets.} to achieve significant improvements on V\&L tasks for which the understanding of text-within-image is crucial.
     \item Overall, \NEWNAME improves SoTA results via fine-tuning on 15+ benchmarks, and we show that it is the first of its kind to simultaneously adapt via multitask fine-tuning to a diverse set of benchmarks without significant performance degradation.
 \end{enumerate}
\section{Related Work}

Similar to large language models such as GPT4~\cite{openai2023gpt4} and PaLM~\cite{chowdhery2022palm}, the benefit of scale has also been observed in recent vision and vision-language models. Flamingo~\cite{alayrac2022flamingo} used a frozen language component and demonstrated the benefit of scaling up this part up to 70B parameters on the few-shot multimodal capabilities, while the vision encoder is fixed with 435M parameters. GIT~\cite{wang2022git}, on the other hand, explored scaling of the vision component up to 4.8B parameter, with a 300M parameter language decoder. \NAME~\cite{pali2} explored jointly scaling the vision and language component, to 4B and 17B, respectively, and showed that scaling both components benefits a wide range of vision-language tasks. All these models took advantage of vision and language unimodal pretrained models as backbones to start multimodal training. Recently, on the vision model side, a vision transformer with 22B parameter has been introduced~\cite{vit-22b}. In this work, we make use of a ViT-22B model specifically tuned for OCR capability to explore scaling Vision-Language models to even larger parameter regime.

As first shown in \cite{brown2020language}, \emph{large} language models are sometimes able to solve new unseen tasks at inference as long as a few examples --or \emph{shots}-- are provided as inputs. This is usually referred to as in-context learning~\cite{dong2022survey}. Follow-up work proposed improved ways to split and prompt the shots, such as Chain of Thought~\cite{wei2022chain} or Least-to-Most prompting~\cite{zhou2022least}. So far, the vast majority of this work has been done in the context of language inputs~\cite{wei2023larger}. In this work, we explore multimodal in-context learning with pairs of images and captions. Our work is aligned in spirit to Flamingo \cite{alayrac2022flamingo} that uses interleaved image text pairs in the same web page and in-context tuning \cite{chen2021meta} during pre-training. We first group the image-text pairs by url and split each group to a ``shots'' set and a ``target'' set. Then we use the few examples in the ``shots'' set as input features to predict the examples in the target set. 

Besides solving vision-language tasks in multiple domains, recent VLMs also attempted solving these tasks at once instead of fine-tuning on each individual benchmark. Unified-IO~\cite{lu2022unified} performed multitask fine-tuning and reported solid results across 16 benchmarks. Spotlight~\cite{li2023spotlight} reported that inside the UI domain, multitask fine-tuning can achieve a performance close to task-specific fine-tuning. In this work, we show that \NEWNAME can be simultaneously fine-tuned with a diverse set of benchmarks in multiple domains without performance degradation. 
\section{Model}
\label{sec:model}
\subsection{Architecture}
\label{section:arch}
The \NEWNAME model architecture follows the encoder-decoder architecture: image(s) are processed by a ViT encoder, with the resulting visual embeddings fed to an encoder-decoder backbone, along with embeddings from additional text input (e.g., question / prefix / prompt). More details are provided in Appendix~\ref{appendix:model_info_and_example}.

\textbf{Visual component}\quad
Our visual backbone is scaled to 22B parameters, as introduced by \cite{vit-22b}, the largest dense ViT model to date.  To equip the model with a variety of complex vision-language tasks, we specifically focus on its OCR capabilities. 
To that end, we incorporate an OCR-based pretraining as follows:
images from the WebLI dataset~\cite{pali2} are annotated with OCR-text detected by GCP Vision API; 
the encoder is then further pre-trained with a mixture of the original JFT-based classification task and a new OCR-based classification task (whether or not a given token occurred in the image according to OCR results).
See Appendix \ref{appendix:model_info_and_example} for additional details on the visual component.
\NEWNAME is designed to take $n>=1$ images as inputs (for few-shot and video understanding), with tasks involving a single image as the $n=1$ case.
For $n>1$, each image is independently processed by the ViT module, and the patch-level embeddings coming out of ViT are flattened and concatenated to form the visual input (See Appendix \ref{appendix:model_info_and_example}).
Note that similar to the single-image case, there is no pooling over the spatial dimension before visual embeddings are aggregated over the temporal dimension.
That is, for an $n$-frame input with $k$-patches per frame, the resulting visual input has $n*k$ tokens.

\textbf{Overall model}\quad
The encoder-decoder backbone is initialized from a variant of the UL2~\cite{tay2023ul2} encoder-decoder model that uses 32B parameters.
The architecture of this variant has 50 layers in both encoder and decoder (up from 32 layers in~\cite{tay2023ul2}), and is pretrained on a mixture of text data similar to~\cite{tay2023ul2}.
The visual embeddings, after going through a projection layer, are concatenated with the token embeddings of the text input, and fed to the encoder-decoder backbone.  Most of the pretraining tasks (with the exception of the masked image token task) predict text-only output from this multimodal input.  The text input to the model typically consists of a prompt that marks what type of task it is
(e.g., \prompt{Generate caption in \lang} for captioning tasks)
and encode necessary textual input for the task
(e.g., \prompt{Answer in \lang: \{question\}} for VQA tasks).
For tasks that need OCR capabilities, we experiment with either relying solely on the text-encoding capabilities of the vision encoder, or optionally including tokens extracted by an upstream OCR system fed as additional text inputs.

\textbf{Few-shot formulation}\quad
In the few-shot setting, for a given {\em target example} the model receives a number of ``labeled'' examples (in the form of additional $\langle$image, text$\rangle$ pairs) that we refer to as \emph{shots}/\emph{exemplars}.
The hypothesis is that information contained in these exemplars provides the model with useful context to generate predictions for the target example.
Formally, the input with $N$ shots is a sequence $(t_1, \dots, t_N, t_T, i_1, \dots, i_N, i_T)$, where $t_1:t_N$ and $i_1:i_N$ are texts and images for the $N$ shots, and $t_T$ and $i_T$ are the text (prompt) and image for the target example.
\NEWNAME processes this input as follows:
all images, including the target one, are first independently processed by the visual encoder, and the resulting patch-level embeddings are flattened and concatenated to form the visual input sequence.
After going through a projection layer, they are concatenated with the text embeddings to form the multimodal input sequence used by the encoder.
We implement additional optimizations including distributing the exemplars between the encoder and the decoder, and an attention re-weighting mechanism (see Appendix~\ref{appendix:image_results}).

\subsection{Pretraining Data and Mixture}
The main pretraining data for our model is based on WebLI~\cite{pali2}, consisting of roughly one billion images with alt-texts from the web and OCR annotations (using the GCP Vision API), covering over 100 languages.
In addition to WebLI $\langle$image, text$\rangle$ pairs, we introduce here {\em Episodic WebLI} data, where each episode corresponds to a set of such pairs.
We aim to have each episode contain loosely related images (i.e., they are clustered according to their URL field), so as to encourage attention among examples in an ``episode''.
We find this new dataset (with 75M episodes and around 400M images in total) important for developing the few-shot capabilities of the model.

The pretraining mixture consists of the following data and objectives:
(i) span corruption on text-only data (15\% of tokens);
(ii) split-captioning on WebLI alt-text data~\cite{simvlm,pali2};
(iii) captioning on CC3M~\cite{cc3m} on native and translated alt-text data (over the same 35 languages covered by \XM~\cite{Thapliyal2022Crossmodal3600AM});
(iv) split-ocr~\cite{kil2022prestu} on WebLI OCR annotations;
(v) visual-question-answering objective over $\langle$image, question, answer$\rangle$ pairs generated using the VQ$^2$A method~\cite{changpinyo-2022-vq2a} over the CC3M training split, over native and translated text (same 35 language pairs);
(vi) visual-question-generation objective, using the same pairs as above;
(vii) visual-question-answering objective over $\langle$image, question, answer$\rangle$ pairs using the Object-Aware method~\cite{piergiovanni-2022-OApretr} (English only);
(viii) captioning on Episodic WebLI examples (target alt-text predicted from the remaining alt-text and images);
(ix) visual-question-answering on 4-pair examples (resembling Episodic WebLI and using VQ$^2$A-CC3M pairs), with the answer target conditioned on the other pairs of $\langle$image, question, answer$\rangle$ data.
(x) pix2struct objective, introduced in ~\cite{lee2022pix2struct}, targeting page layout and structure using screenshot images paired with DOM-tree representations of html pages.
(xi) Captioning on short video data, using the VTP data~\cite{alayrac2022flamingo} (using four frames per video).
(xii) object-detection objective on WebLI data, whereby an OWL-ViT model~\cite{owlvit} (L/14) is used to  annotate WebLI images, resulting in hundreds of pseudo object labels and bounding boxes per image.
(xiii) image-token prediction objective, whereby we tokenize WebLI images (\imres{256} resolution) using a ViT-VQGAN~\cite{yu2021vector} model with patch size \imres{16} ($256$ tokens per image); this objective is framed as a 2D masked-token task (i.e., fill-in the missing grid pieces, with the corresponding image pixels also masked).
Note that the image-token prediction objective is added mainly as a condition to check whether it adversarially impacts the performance on language-output tasks; our ablation experiments show that is does not.

\subsection{Training Stages}
Our model is trained in two stages.
In stage 1, the visual encoder (after mixed-objective training) is kept frozen, while the rest of the parameters are trained on a total of 2.2B examples at the base resolution \imres{224} (native to ViT-22B), using the entire mixture.
In stage 2, it continues training using only the OCR-related objectives (pix2struct and split-ocr) plus the object detection objective; this is done in several substages, during which image resolution is gradually increased to \imres{448}, \imres{672} and finally \imres{756}.

\section{Experiments}

\subsection{Image Captioning and Visual Question Answering}
\label{sec:exp:finetuning}

Our results demonstrate that the larger capacity in \NEWNAME scales well in both its vision and language components, and it is particularly beneficial for more challenging scene-text and document understanding tasks. Our model outperforms the SOTA on diverse vision-language tasks, with significant margins in some cases.

\textbf{Benchmark datasets}\quad The Image Captioning and VQA benchmarks used for evaluation is summarized in Appendix~\ref{appendix:image_results}, including 6  Image Captioning benchmarks (COCO (Karpathy split~\cite{cocokarp}), NoCaps~\cite{nocaps}, TextCaps~\cite{textcaps}, VizWiz-Cap~\cite{gurari2020captioning}, Screen2Words~\cite{screen2words}, Widget-Cap~\cite{li-etal-2020-widget}) and 13 VQA benchmarks (VQAv2~\cite{vqav2}, OKVQA~\cite{marino2019okvqa}, TallyQA~\cite{acharya2019tallyqa}, TextVQA~\cite{textvqa}, VizWiz-VQA~\cite{gurari2018vizwiz}, STVQA~\cite{ST-VQA}, OCRVQA~\cite{mishraICDAR19}, InfographicVQA~\cite{mathew2022infographicvqa}, DocVQA~\cite{mathew2021docvqa}, AI2D~\cite{kembhavi2016diagram} ChartQA~\cite{masry2022chartqa}, OVEN~\cite{hu2023open}, InfoSeek~\cite{chen2023can}). These tasks span a wide range of visual domains, from natural images, illustrations to documents and user interfaces (UIs). We also include results of multilingual captioning on \XM in Appendix~\ref{appendix:image_results}.

\subsubsection{Per-task fine-tuning results}
\textbf{Experimental setup}\quad We fine-tune \NEWNAME with frozen ViT-22B; the learning rate follows a linear decay from initial value 1e-4 for all fine-tuning experiments. See Appendix~\ref{appendix:image_results} for more details.

\begin{table*}[ht!]
\centering
\resizebox{\linewidth}{!}{%
\begin{tabular}{lcccccccccccc}
\toprule
& COCO && \multicolumn{2}{c}{NoCaps} && \multicolumn{2}{c}{VQAv2} && OKVQA && \multicolumn{2}{c}{TallyQA} \\
\cmidrule{2-2} \cmidrule{4-5} \cmidrule{7-8} \cmidrule{10-10} \cmidrule{12-13}
Model & Karp.-test & & val & test & & test-dev & test-std && val && simple & complex \\
\midrule
GIT2~\cite{wang2022git} (5.1B) & 145.0 & & 126.9 & \textbf{124.8} && 81.74 & 81.92 && - && - & -\\
Flamingo~\cite{alayrac2022flamingo} (80B) & 138.1 & & - & - && 82.0 & 82.1 && 57.8$^*$ && - & -\\
BEiT-3~\cite{beit3} (1.9B) & 147.6 && - & - && 84.2 & 84.0 && - && - & -\\
PaLM-E~\cite{driess2023palme} (562B) & 138.7 && - & - && 80.0 & - && \textbf{66.1} & - & - \\
MoVie~\cite{movie2021} & - && - & - && 69.26 & - && - && 74.9 & 56.8 \\
\NAME~\cite{pali2}(17B) & 149.1 & & \textbf{127.0} & 124.4 & & 84.3 & 84.3 && 64.5 && 
81.7 & 70.9 \\
\midrule
\NEWNAME(55B) & \textbf{149.2} && 126.3 & 124.3 && \textbf{86.0} & \textbf{86.1} && \textbf{66.1} && \textbf{86.0} & \textbf{75.6}\\
\bottomrule
\end{tabular}}
\caption{Results on COCO Captions (Karpathy split), NoCaps, VQAv2~\cite{vqav2}, OKVQA~\cite{marino2019okvqa}, and TallyQA~\cite{acharya2019tallyqa} with end-to-end modeling without OCR pipeline input (``simple'' and ``complex'' are test subsplits).
}
\label{table:captioning}
\end{table*}

\begin{table*}[ht!]
\centering
\resizebox{\linewidth}{!}{%
\begin{tabular}{lc@{\hspace{0.1cm}}c@{\hspace{0.1cm}}c@{\hspace{0.08cm}}c@{\hspace{0.08cm}}c@{\hspace{0.08cm}}c@{\hspace{0.08cm}}c@{\hspace{0.08cm}}c@{\hspace{0.08cm}}c@{\hspace{0.08cm}}c@{\hspace{0.08cm}}c@{\hspace{0.08cm}}c@{\hspace{0.08cm}}c@{\hspace{0.08cm}}c}
\toprule
& Text & VizWiz & Text & VizWiz & ST & OCR & Info & Doc & \multirow{2}{*}{AI2D} & Chart & Screen2 & Widget & \multirow{2}{*}{OVEN} & Info \\
Model & Caps & Cap & VQA & VQA & VQA & VQA & VQA & VQA & & QA & Words & Cap & &Seek\\
\midrule
\multicolumn{4}{l}{\textit{\textbf{with} OCR pipeline input}} &&&&&&&&&& \\
\midrule
\multirow{2}{*}{SoTA}
& 160.4 & 124.7 & 73.67 & 73.3 & 79.9 & 67.5 & 47.4 & 84.7 & 38.5 & 45.5 & - & - & - & - \\
&
\small{\cite{pali2}} &
\small{\cite{pali2}} &
\small{\cite{mia_textvqa}} & \small{\cite{pali2}} & \small{\cite{pali2}} & \small{\cite{biten2022latr}} & \small{\cite{tang2022unifying}} & \small{\cite{tang2022unifying}} & \small{\cite{kembhavi2016diagram}} & \small{\cite{masry2022chartqa}} & - & - & - & - \\
\NEWNAME & \textbf{163.7} & \textbf{125.7} & \textbf{80.78} & \textbf{74.6} & \textbf{84.5} & \textbf{77.3} & \textbf{54.8} & \textbf{86.8} & \textbf{81.4} & \textbf{72.3} & - & - & - & - \\
\midrule
\multicolumn{4}{l}{\textit{\textbf{without} OCR pipeline input}} \\
\midrule
\multirow{2}{*}{SoTA}
& 145.0 & 120.8 & 67.27 & 70.7 & 75.8 & 71.3 & 40.0 & 76.6 & 42.1 & 70.5 & 109.4 & 141.8 & 20.0 & 17.7 \\
& \small{\cite{wang2022git}} & \small{\cite{wang2022git}} & \small{\cite{wang2022git}} & \small{\cite{pali2}} & \small{\cite{wang2022git}} & \small{\cite{lee2022pix2struct}} & \small{\cite{lee2022pix2struct}} & \small{\cite{lee2022pix2struct}} & \small{\cite{lee2022pix2struct}} & \small{\cite{deplot}} &
\small{\cite{lee2022pix2struct}} & \small{\cite{li2023spotlight}} &
\small{\cite{hu2023open}} & \small{\cite{chen2023can}} \\
\NEWNAME & \textbf{147.0} & \textbf{122.7} & \textbf{71.44} & \textbf{70.9} & \textbf{79.9} & \textbf{75.0} & \textbf{49.2} & \textbf{80.0} & \textbf{81.2} & \textbf{70.9} & \textbf{127.9} & \textbf{153.0} & \textbf{23.1} & \textbf{21.8} \\
\bottomrule
\end{tabular}}
\caption{Results on benchmarks more focused on text understanding capabilities.
For OVEN~\cite{hu2023open} \& InfoSeek~\cite{chen2023can}, we follow the proposed \imres{224} resolution settings for fair comparison.
}
\label{table:scene-text-like}
\end{table*}

First, we present benchmarks results for the condition where external OCR systems are not used (Table \ref{table:captioning}, see Appendix~\ref{appendix:image_results} for an extended table.).
The trend is that \NEWNAME matches or improves SoTA results on these benchmarks, with a particularly significant improvement on the TallyQA benchmark over MoVie~\cite{movie2021} (specialized counting model), at +11.1 for simple counting questions (e.g., ``how many giraffes'') and +18.8 for complex counting questions (e.g., ``how many giraffes are drinking water''); there are significant improvements over PaLI~\cite{pali2} as well, indicating that scale plays an important role in the ability of such models to perform counting tasks.
We additionally note the state-of-the-art result on VQAv2 at 86.1 accuracy, achieved with an open-vocabulary generative approach, and the performance on OKVQA at 66.1 accuracy, matching the much-larger PaLM-E~\cite{driess2023palme} model performance.

Next, we examine text-heavy V\&L benchmarks, for which upstream OCR systems can be used to improve performance.
As shown in Table~\ref{table:scene-text-like},
\NEWNAME improves SoTA for all Captioning and VQA benchmarks across the board, either without  or with additional OCR input (using GCP Vision API).
For instance, a significant jump of +42.9 points is observed on AI2D\footnote{As with all the other benchmarks, our training examples are carefully deduped to exclude images occurring in these benchmarks, including AI2D. Such results, therefore, are \emph{not} attributable to train-test data leakage.}, a multiple-choice benchmark where choices are provided along with each question. Being able to have the text choices as input benefits \NEWNAME compared with the previous SoTA Pix2Struct~\cite{lee2022pix2struct} which has to render the text on the image, but this does not explain all the improvements. In a question-only configuration (no answer choice present), \NEWNAME achieves 46.3 on AI2D, more than 4 points higher than Pix2Struct's result.

In general, having access to OCR texts extracted by an external OCR pipeline boosts performance. Still, for several benchmarks (e.g., AI2D, ChartQA, OCRVQA and Widget-Cap), \NEWNAME's end-to-end performance when using its intrinsic OCR capability is close to that leveraging additional OCR input. A common feature for these benchmarks is that they have well-oriented text -- diagrams, charts, book covers or user interfaces, with reasonably large font size at \imres{756} resolution. For tasks involving scene text in natural images (TextCaps, TextVQA, STVQA) or very high density of small texts (DocVQA, InfoVQA), results still highlight clear benefits when utilizing an external OCR model.

\subsubsection{Multitask Fine-tuning}

We simultaneously fine-tune and evaluate the pretrained checkpoints on multiple benchmarks belonging to the same category. We deduplicated every training set over the test sets of every task in the mixture to prevent the leakage of any test-set examples into the mixed training set.
This is useful as it leads to a single fine-tuned model that performs all the tasks, rather than having to fine-tune each task separately.
We performed such multitask fine-tuning on all Image Captioning benchmarks and most VQA benchmarks, respectively.

Table~\ref{table:multitask_cap} shows the multitask fine-tuning result for captioning tasks. The performance on COCO is slightly decreased in the multitask setting, which is likely a result of 
this task needing longer training to converge.
For Screen2Words, having the smallest train and dev/test sets could be responsible for the performance fluctuation. Notably, VizWiz-Cap and Widget-Cap shows improved performance from multitask fine-tuning. Overall, the average performance decreases by 1.4 points (0.2 excluding Screen2Words) with multitask fine-tuning, while offering the clear advantage of having a single checkpoint to perform all these tasks.
Appendix~\ref{appendix:image_results} shows similar results for VQA tasks.
We consider this outcome a positive result that establishes the on-par performance between multitask fine-tuning and single-task fine-tuning for diverse benchmarks, in contrast with previous work which argued a gap between single-task and multitask fine-tuning~\cite{lu2022unified}, or demonstrated little gap over benchmarks from the same domain~\cite{li2023spotlight}.

\begin{table*}[ht!]
\centering
\resizebox{0.9\linewidth}{!}{%
\begin{tabular}{lc@{\hspace{0.2cm}}c@{\hspace{0.2cm}}c@{\hspace{0.2cm}}c@{\hspace{0.2cm}}c@{\hspace{0.2cm}}c@{\hspace{0.2cm}}c}
\toprule
& \multirow{2}{*}{COCO} & \multirow{2}{*}{NoCaps} & Text & VizWiz & Screen2 & Widget & Avg. \\
Method &&& Caps & Cap & Words & Cap\\
\midrule
Split & Karp.-test & val & val & test-dev & test & test & -\\
\midrule
SOTA (Single-task FT) & 149.1 & \textbf{127.0} & 148.6 & 119.4 & 109.4 & 136.7\\
\midrule 
\NEWNAME Single-task FT & \textbf{149.2} & 126.3 & 150.8 & 123.1 & \textbf{127.9} & 153.2 & - \\
\NEWNAME Multitask FT & 147.3 & 125.6 & \textbf{154.6} & \textbf{124.2} & 120.6 & \textbf{153.7} & - \\
Multitask (+/-) & \red{-1.9} & \red{-0.7} & \textcolor{teal}{+3.8} & \textcolor{teal}{+1.1} & \red{-7.3}$^*$ & \textcolor{teal}{+0.5} & \red{-1.4} (\red{-0.2} w/o ``*'') \\
\bottomrule
\end{tabular}}
\caption{Scores from multitask fine-tuning compared with those from single-task fine-tuning for Image Captioning. Validation or test-dev set numbers are reported for some tasks.}
\label{table:multitask_cap}
\end{table*}

\subsubsection{Few-shot Evaluation}

We fine-tuned the \NEWNAME model on a mixture of few-shot tasks. The few-shot mixture contains Episodic mixtures, (Non-Episodic) Webli and (Non-Episodic) CC3M data.
Note that all of these datasets were already used in previous stages of training, but with lower mixture proportions.
During pre-training, we only use up to 4 shots, with both encoder and decoder shots (see Appendix~\ref{appendix:image_results}).
For fine-tuning, we use up to 8 encoder shots and do not use decoder shots.

We evaluate the few-shot performance on COCO caption (Karpathy test split~\cite{cocokarp}), and \XM \cite{Thapliyal2022Crossmodal3600AM} datasets.
For each task, we first create a ``shots pool'' with 256 examples that are randomly selected from the task's training set.
As the \XM benchmark does not come with a training set, we use Google Translate API to enhance the COCO Karpathy training set with captions in the 35 languages represented in \XM.
Then, for each test data point, we randomly pick $N$ shots from the pool as the actual few-shot examples. Following \cite{alayrac2022flamingo}, we also evaluate on 2 text-only shots settings where only the textual part of 2 randomly sampled few-shot examples are used.

Table~\ref{table:fewshot_eval} reports the few-shot captioning performance on English and multilingual captioning, as well as few-shot VQA performance on VQAv2.
\NEWNAME achieves SOTA few-shot results on COCO with both 4 shots and 32 shots; it outperforms previous SOTA by +4.4 CIDEr points for 4-shot, suggesting a strong ability to efficiently gather hints from few examples. 
We also report few-shot CIDEr scores averaged over 35 languages using \XM, demonstrating \NEWNAME's multilingual capabilities. Meanwhile, although \NEWNAME also performs decently on VQAv2, the gap behind the SoTA Flamingo model~\cite{alayrac2022flamingo} (which freezes the language backbone) may be the result of losing some of the few-shot text-only QA capability by fine-tuning the language backbone, which supports the hypothesis regarding the tension between few-shot and fine-tuning abilities. 

\begin{table*}[ht!]
\centering
\resizebox{\linewidth}{!}{%
\begin{tabular}{lcccccccc}
\toprule
 & \multicolumn{2}{c}{COCO Captions} && \multicolumn{2}{c}{\XM Cap. (35-lang avg.)} && \multicolumn{2}{c}{VQAv2}  \\ 
 \cmidrule{2-3} \cmidrule{5-6} \cmidrule{8-9}
 Method & 4 shots & 32 shots && 4 shots & 32 shots && 4 shots & 32 shots \\
 \midrule
Prev. SoTA \cite{alayrac2022flamingo} & 103.2 & 113.8 && \multicolumn{2}{c}{N/A (53.6 w/ fine-tune~\cite{pali2})} && \textbf{63.1} & \textbf{67.6} \\
\NEWNAME   & \textbf{107.6} & \textbf{114.5} && 45.1 & 47.1 && 56.9 & 57.1  \\ \bottomrule
\end{tabular}}
\caption{Few-shot performance of the \NEWNAME model (multilingual captioning for \XM).}
\label{table:fewshot_eval}
\end{table*}

\subsection{Video Captioning and Question Answering}

We fine-tune and evaluate the \NEWNAME model on 4 video captioning (MSR-VTT~\cite{xu2016msr}, VATEX~\cite{wang2019vatex}, ActivityNet Captions~\cite{krishna2017dense}, Spoken Moments in Time~\cite{monfort2021spoken}) and 3 video question answering benchmarks (NExT-QA~\cite{xiao2021next}, MSR-VTT-QA~\cite{xu2017video}, ActivityNet-QA~\cite{yu2019activitynet}).
A brief description of each benchmark and clarifications on their usage are provided in Appendix~\ref{appendix:video_results}.

\begin{table}[t]
\centering
\resizebox{\linewidth}{!}{%
\begin{tabular}{lccccccc}
\toprule
& \multicolumn{2}{c}{MSR-VTT} & \multicolumn{2}{c}{Activity-Net} & VATEX & SMIT & NExT-QA \\
\cmidrule{2-8}
Method&Cap.~\cite{xu2016msr} & QA~\cite{xu2017video} &Cap.~\cite{krishna2017dense} &QA~\cite{yu2019activitynet} 
&Cap.~\cite{wang2019vatex} &Cap.~\cite{monfort2021spoken} &QA~\cite{xiao2021next} \\
\midrule
Prior SOTA & 75.9 & \textbf{47.4} & 52.5 & 44.7 & 94.0$^\dagger$ & 28.1$^\ddagger$ &  33.5$^\mathsection$ \\
& \sotamodel{GIT2~\cite{wang2022git}} &
\sotamodel{Flamingo~\cite{alayrac2022flamingo}} &
\sotamodel{PDVC~\cite{wang2021end}} &
\sotamodel{VINDLU~\cite{cheng2022vindlu}} &
\sotamodel{GIT2~\cite{wang2022git}} &
\sotamodel{MV-GPT~\cite{seo2022end}} &
\sotamodel{Flamingo 32shot~\cite{alayrac2022flamingo}}
\\
\midrule
\NEWNAME (8fr) & 74.6 & 46.9 & 49.0 & 48.4 & 66.0 & 42.5 & 37.0\\
\NEWNAME (16fr) & \textbf{76.8} & \textbf{47.1} & \textbf{54.9} & \textbf{49.4} & 69.3 & \textbf{43.5} & \textbf{38.3} \\
\bottomrule
\end{tabular}}
\vspace{1pt}
\caption{Results for Video Captioning and Video-QA using 8 frames (8fr) or 16 frames (16fr).  $\dagger$GIT2 uses Self-Critical Sequence Training to directly optimize the CIDEr metric for VATEX. $\ddagger$SMIT has not been used for video captioning before, we apply MV-GPT~\cite{seo2022end} and report results on the test set. $\mathsection$Numbers were obtained using 32-shot; since Flamingo 32-shot outperforms fine-tuning SOTA on this open-ended QA task, they did not conduct further fine-tuning experiments for this task.
}
\label{table:videoresults}
\end{table}

\textbf{Experimental setup}\quad
We fine-tune our model (with base resolution \imres{224}) for each task separately, use the validation split for early stopping, and report performance on the test split.
We use a learning rate of $10^{-4}$ for all tasks, and do not adapt any hyperparameters for specific tasks. 
Frames are sampled using a fixed temporal stride for each dataset (determined based on the video length distribution in that dataset such that the product of the number of frames and stride is larger than the total number of frames for half of the videos), and we experimented with including up to 8 or 16 frames per video.  We did not include pooling over the spatial dimension; embeddings for \imres{16} patches per frame are provided as visual input to the multimodal encoder.

\textbf{Results}\quad
We report CIDEr score for the video captioning tasks.
Video QA tasks are treated as open-ended generation tasks;
we report full-string accuracy (for MSR-VTT-QA and ActivityNet-QA) and WUPS metrics (NExT-QA) in~\cite{amultiworld,xiao2021next}.
As shown in Table \ref{table:videoresults}, the 16-frames version has an edge over the 8-frame version, sometimes with a significant margin (e.g., close to a 6 point increase in CIDEr score for ActivityNet-Captions).
More importantly, while \NEWNAME pretraining was dominated by image-text tasks, we were able to achieve new SOTA performance for 5 out of 7 tasks\footnote{As noted in Table~\ref{table:videoresults}, current SOTA on NExT-QA for the open-ended QA task was achieved by Flamingo 32-shot, which had outperformed prior fine-tuning SOTA.  To the best of our knowledge, \NEWNAME performance on this task does outperform existing published fine-tuning performances, with the caveat that we do not have information on what Flamingo fine-tuning would have achieved on this task.}, and performed very close to prior SOTA on MSR-VTT-QA (47.1 vs 47.4).

\subsection{Image classification}

To test image classification capabilities we fine-tuned \NEWNAME and models from \cite{pali2} on ImageNet~\cite{deng2009imagenet} and evaluated the resulting model on \mbox{ImageNet-REAL}~\cite{imagenet_real} and out-of-distribution datasets: \mbox{ImageNet-R}~\cite{hendrycks2021many}, \mbox{ImageNet-A}~\cite{hendrycks2021nae}, \mbox{ImageNet-Sketch}~\cite{wang2019learning}, \mbox{ImageNet-v2}~\cite{recht2019imagenet}.
We used the model from the first training stage (at resolution 224) and the one from the last training stage (at resolution 756).
We used the same training hyperparameters for all of runs (selected without any hyperparameter tuning; mode details in Appendix \ref{appendix:img_classification_results}).

The results can be seen in Table~\ref{table:imagenet_training}. 
We compare the results to generative model with open vocab -- GIT2~\cite{wang2022git} (using 384 image resolution), which is the current SOTA for full fine-tuning on ImageNet. \NEWNAME achieves SOTA results for generative models on Imagenet, and other datasets. We also performed zero-shot evaluation for \NEWNAME and the results can be found in Appendix \ref{appendix:img_classification_results}.

\begin{table*}[ht!]
\centering
\resizebox{0.95\linewidth}{!}{
\begin{tabular}{lccccccc}
\toprule
Model (resolution) & INet~\cite{deng2009imagenet} & REAL~\cite{imagenet_real} & INet-R~\cite{hendrycks2021many} &  INet-A~\cite{hendrycks2021nae}  & INet-Sketch~\cite{wang2019learning} &  INet-v2~\cite{recht2019imagenet} \\
\midrule
GIT2~\cite{wang2022git} (384)     & \textbf{89.22}  &   - &   -       &  -      & -    & -     \\
PaLI-17B~\cite{pali2} (224) & 86.13  &   88.84 &   78.21       &  50.00      & 71.21    & 78.91     \\
\midrule
\NEWNAME (224)  & 88.22 & 90.36 & 77.66  &  55.97  &  72.56   &  81.42     \\
\NEWNAME (756) & \textbf{89.19} &  \textbf{90.98}   &  \textbf{80.06}   &  \textbf{72.57}    &  \textbf{73.37}    & \textbf{83.66} \\
\bottomrule
\end{tabular}}
\caption{Classification accuracy (top-1) fine-tuned on Imagenet~\cite{deng2009imagenet}.} 
\label{table:imagenet_training}
\end{table*}

\begin{figure}[ht]
    \centering
    \includegraphics[width=0.99\linewidth]{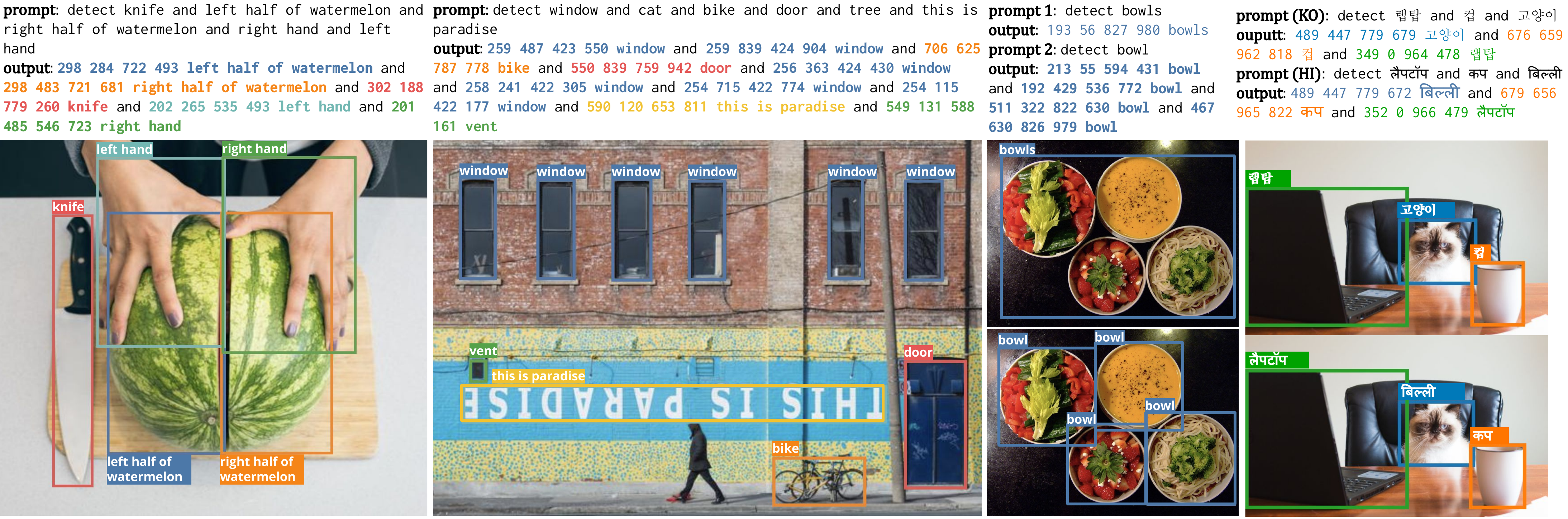}
    \footnotesize{\textit{Credits: Watermelon/Cat; Sarah Pflug (burst), Bowls; ariesandrea (flickr), Wall; Matthew Henry (burst)}}
    \caption{Examples demonstrating multilingual, OCR and other capabilities transferred to detection.}
    \label{fig:det_examples}
\end{figure}

\subsection{Object Detection}
Object detection can be easily formulated in our model as shown in pix2seq~\cite{chen2022pix2seq}, The dataset mix used for pre-training is presented in Sec.~\ref{sec:model}; detection data was included up to and including the stage using resolution 672, after which a separate detection-specific model was fine-tuned on detection data. Before detection-specific tuning, LVIS~\cite{gupta2019lvis} \& COCO labels were removed from all detection training datasets, allowing zero-shot evaluation on LVIS.

Bounding box mean AP on LVIS is shown in Table ~\ref{table:detection_results}, including zero-shot performance; the detection-tuned model reaches an AP of 31 in general, and 31.4 on rare classes, and about 12 for both in zero-shot. Performance on rare classes was on par with performance on common classes, a difficult feat traditionally accomplished by complicated sampling schedules and augmentations.
In our set up, it is directly enabled by \NEWNAME's diverse training mix.
This could likely be further improved with investment in fine-tuning e.g. using noise-augmentation methods from pix2seq~\cite{chen2022pix2seq}, or a further stage of high-resolution, LVIS only training. 
Qualitatively, we observe emergence of many interesting phenomena enabled by co-training with non-detection tasks; for example, multilingual detection, OCR bounding boxes and longer descriptions, none of which are included in detection training, are often handled well by \NEWNAME. Additional results and information can be found in Appendix~\ref{appendix:detection_information}.
\begin{table*}[h]
\centering
\begin{tabular}{@{}lcc@{}}
\toprule
                                     & LVIS AP & LVIS AP\textsubscript{Rare} \\ \midrule

ViLD~\cite{vild} (tuned on non-rare LVIS) &  29.3  & 26.3                                \\ 
Region-CLIP~\cite{regionclip} (tuned on non-rare LVIS) &  32.3  & 22.0                                \\ 
OwLViT-L/16~\cite{owlvit} (tuned on non-rare LVIS) &  34.7  & 25.6                                \\ 
OwLViT-L/16~\cite{owlvit} (with Object365 and VG datasets)               & 34.6  & 31.2                                             \\

\midrule
\NEWNAME (Zeroshot)                             & 12.36 & 12.16                                      \\
 \NEWNAME (Detection-tuned)                     & 30.64 & 31.42                                      \\
\bottomrule
\end{tabular}

\caption{\NEWNAME object detection results on LVIS. The diverse pre-training mix enables parity performance between LVIS rare and common classes. Other related approaches are shown for context, but are not directly comparable.}
\label{table:detection_results}
\end{table*}
\section{Model Fairness, Biases, and Other Potential Issues}
Large models, if left unchecked, have the potential to inflict harm on society -- such as amplifying biases ~\cite{hendricks2018women,caliskan2017semantics,zhao2017men,wang2020towards}, causing  disparities~\cite{zhao2017men,buolamwini2018gender,deuschel2020uncovering}, or encoding narrow cultural perspectives \cite{hutchinson2022underspecification,de2019does}.
Hence, evaluating \NEWNAME for such potential issues is important.
We focus our RAI evaluation on three parts: (1) harmful  associations, such as toxicity and profanity, (2) demographic parity in the model's output, such as encoding societal stereotypes/biases, and (3) performance disparity across  subgroups. This breakdown follows earlier works in the literature, such as~\cite{goyal2022fairness}. 

\textbf{Toxicity / profanity.}\quad
 We estimate the level of toxicity and profanity in the generated captions, including when disaggregated across subgroups. We use the FairFace dataset~\cite{karkkainen2021fairface} that comprises of images of people with ground-truth attributes: gender presentation, age and ethnicity.
We generate captions and use the Perspective API~\cite{lees2022new} (threshold $>0.8$) to measure toxicity and profanity. Table \ref{table:toxicity} summarizes the results; we observe a low level of toxicity/profanity across all slices. Tables \ref{table:toxicity_race_appendix} and \ref{table:toxicity_age_appendix} provide a detailed breakdown of toxicity/profanity results for all subgroups in FairFace dataset. In Tables~\ref{table:toxicity_miap_gender_appendix} and~\ref{table:toxicity_miap_age_appendix}, we report similar results in the MIAP~\cite{miap_aies} dataset, disaggregated by perceived gender and age. 

\begin{table*}[ht!]
\centering\footnotesize
\resizebox{0.95\linewidth}{!}{%
\begin{tabular}{l|cc|ccc|ccc|c}
\toprule
 & \multicolumn{2}{c}{Gender} & \multicolumn{3}{c}{Ethnicity} & \multicolumn{3}{c}{Age} & \\
 &  Lowest & Highest & Lowest   & Median & Highest & Lowest & Median & Highest & \bf Overall\\
\midrule
\bf Toxicity &0.14\% & 0.19\% &  0.00\% & 0.13\% & 0.39\% & 0.00\% & 0.17\% & 0.31\% & \bf 0.01\%\\
\bf Profanity  & 0.00\% & 0.02\% & 0.00\% & 0.00\% & 0.05\% & 0.00\% & 0.00\% & 0.03\% & \bf 0.00\%\\
\bottomrule
\end{tabular}}
\caption{Average toxicity/profanity in the captions generated by \NEWNAME on FairFace dataset.} 
\label{table:toxicity}
\end{table*}

\begin{table*}[ht!]
\centering\footnotesize
\begin{tabular}{l|ccc|ccc}
\toprule
Ethnicity & \multicolumn{3}{c}{Toxicity} & \multicolumn{3}{c}{Profanity} \\
\midrule
& $< 0.2$   & $0.2-0.8$ & $>0.8$    & $< 0.2$   & $0.2-0.8$ & $>0.8$ \\ \midrule
Middle Eastern   &64.24\%   &35.76\%   &0.00\%   &94.87\%   &5.13\%   &0.00\%   \\
Black  &59.47\%   &40.40\%   &0.13\%   &92.67\%   &7.33\%   &0.00\%   \\
Indian     &63.86\%   &36.07\%   &0.07\%   &94.39\%   &5.61\%   &0.00\% \\
Hispanic  & 61.09\%  & 38.79\%  & 0.12\%  &94.45\%   & 5.55\%  & 0.00\%  \\
White  &62.45\%   &37.16\%   &0.39\%   &92.85\%   & 7.10\%  & 0.05\%  \\
Southeast Asian  &63.18\%   &36.61\%   &0.21\%   &93.57\%   & 6.43\%  &0.00\%   \\
East Asian   &63.15\%   &36.72\%   &0.13\%   & 91.55\%  & 8.45\%  &  0.00\% \\
\bottomrule
\end{tabular}
\caption{Distribution of the predicted toxicity/profanity for the captions generated by \NEWNAME on FairFace dataset disaggregated by ethnicity.}
\label{table:toxicity_race_appendix}
\end{table*}

\begin{table*}[ht!]
\centering\footnotesize
\begin{tabular}{l|ccc|ccc}
\toprule
Age & \multicolumn{3}{c}{Toxicity} & \multicolumn{3}{c}{Profanity} \\
\midrule
& $< 0.2$   & $0.2-0.8$ & $>0.8$    & $< 0.2$   & $0.2-0.8$ & $>0.8$ \\ \midrule

< 19     &58.78\%&40.00\%&0.22\%&89.71\%&10.29\%&0.00\%\\

20 - 29  &63.01\%&36.86\%&0.12\%&93.24\%&6.73\%&0.03\% \\

30 - 39  &63.13\%&36.70\%&0.17\%&95.41\%&4.59\%&0.00\%  \\

40 - 49   &63.62\%&36.31\%&0.07\%&95.27\%&4.73\%&0.00\%  \\

50 - 59 &65.87\%&33.88\%&0.25\%&96.48\%&3.52\%&0.00\%\\

60 - 69 &65.31\%&34.38\%&0.31\%&95.95\%&4.05\%&0.00\% \\

$>70$ &66.10\%&33.90\%&0.00\%&92.37\%&7.63\%&0.00\% \\
\bottomrule
\end{tabular}
\caption{Distribution of the predicted toxicity/profanity for the captions generated by \NEWNAME on FairFace dataset disaggregated by age.}
\label{table:toxicity_age_appendix}
\end{table*}

\begin{table*}[ht!]
\centering\footnotesize
\begin{tabular}{l|ccc|ccc}
\toprule
Perceived Gender & \multicolumn{3}{c}{Toxicity} & \multicolumn{3}{c}{Profanity} \\
\midrule
& $< 0.2$   & $0.2-0.8$ & $>0.8$    & $< 0.2$   & $0.2-0.8$ & $>0.8$ \\ \midrule
Predominantly Feminine   &53.98\%   &45.93\%   &0.09\%   &90.55\%   &9.39\%   &0.07\%   \\
Predominantly Masculine  &70.76\%   &29.17\%   &0.06\%   &94.97\%   &5.01\%   &0.01\%   \\
\bottomrule
\end{tabular}
\caption{Distribution of the predicted toxicity/profanity for the captions generated by \NEWNAME on MIAP dataset disaggregated by perceived gender.}
\label{table:toxicity_miap_gender_appendix}
\end{table*}

\begin{table*}[ht!]
\centering\footnotesize
\begin{tabular}{l|ccc|ccc}
\toprule
Age Bucket & \multicolumn{3}{c}{Toxicity} & \multicolumn{3}{c}{Profanity} \\
\midrule
& $< 0.2$   & $0.2-0.8$ & $>0.8$    & $< 0.2$   & $0.2-0.8$ & $>0.8$ \\ \midrule
0-2 yrs   &28.00\%   &72.00\%   &0.00\%   &69.90\%   &30.10\%   &0.00\%   \\
3-19 yrs  &49.96\%   &49.96\%   &0.07\%   &91.46\%   &8.54\%   &0.00\%   \\
20-59 yrs     &66.27\%   &33.68\%   &0.05\%   &93.42\%   &6.55\%   &0.03\% \\
> 60 yrs  & 65.46\%  & 34.54\%  & 0.00\%  &96.39\%   & 3.61\%  & 0.00\%  \\
\bottomrule
\end{tabular}
\caption{Distribution of the predicted toxicity/profanity for the captions generated by \NEWNAME on MIAP dataset disaggregated by age bucket.}
\label{table:toxicity_miap_age_appendix}
\end{table*}

\textbf{Bias / Demographic Parity.}\quad
 We  estimate the level of demographic parity (DP) \cite{dwork2012fairness} in \NEWNAME with respect to gender and occupation. 
To estimate the level of demographic parity (DP) in the model's output, we feed an image into \NEWNAME with the chosen occupation title as a prefix and record the average log-perplexity score of the captions generated by the model.
To ensure that any observed parity would likely reflect unintended biases in the model itself as opposed to the evaluation dataset,   we use CelebA \cite{liu2015faceattributes} that contains celebrity images with gender presentation annotation. Our assumption is that many occupations reflecting societal stereotypes, such as secretaries and plumbers, are quite rare in the CelebA dataset so disparities in output may reflect what is encoded in the model itself.  The list of occupations is compiled based on~\cite{rae2021scaling} and the US job statistics report in~\cite{rudinger-EtAl:2018:N18}. 

\begin{figure}[ht!]
    \hspace*{7pt}\includegraphics[width=0.94\columnwidth]{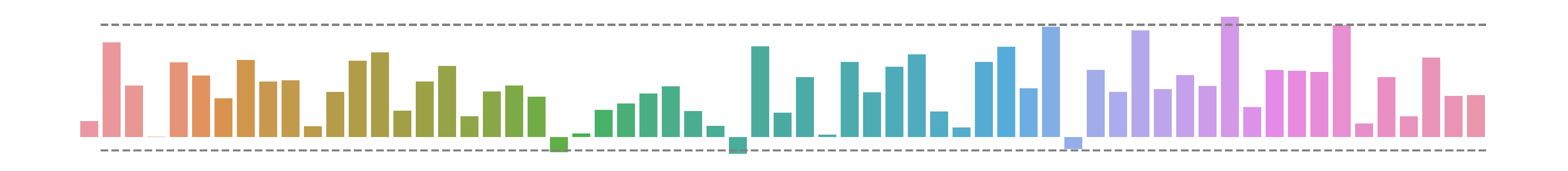}
    \includegraphics[width=\columnwidth]{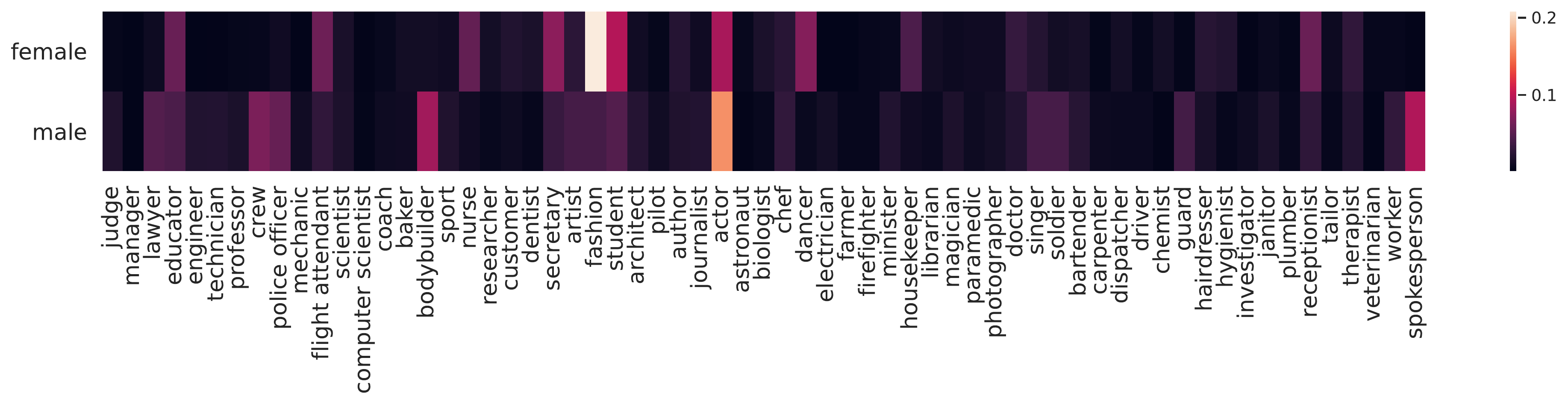}
    \caption{
    {\sc top:} Level of demographic parity (DP) in \NEWNAME's output for CelebA images between women and men. Values close to zero indicate absence of bias.  {\sc bottom:} \emph{Absolute} Pearson correlation coefficients between gender presentation and occupations in WebLI.}
    \label{fig:representation_webli}
\end{figure}

Figure \ref{fig:representation_webli} ({\sc top}) summarizes the overall results. First, \NEWNAME tends to assign a higher log-perplexity score to women than men across most occupations; i.e.  men are predicted to be more likely to hold such occupations. Second, 
\NEWNAME assigns a higher likelihood for a woman to be (`secretary' \& `actor') and  a higher likelihood for a man to be (`guard' \& `plumber') at the 95\% confidence level. Figure \ref{fig:representation_webli} ({\sc bottom}) displays the corresponding correlations between perceived gender presentation and occupations within the WebLI dataset, where we use the Pearson correlation coefficient by treating each label as a binary random variable and noting that for binary random variables, zero correlation implies full independence. All absolute correlation coefficients in the data are $<0.2$ with 99\% of them being $<0.1$.

\textbf{Performance Disparity.}\quad
We present here an evaluation of how well \NEWNAME performs across different subgroups using the MIAP~\cite{miap_aies} dataset. For images containing exactly a single individual, we query \NEWNAME with the question: ``Is there a person in this image?'' and evaluate the accuracy of its response. Note that there are no false positives in this evaluation. Table~\ref{table:miap_rai} summarizes the results. We observe that \NEWNAME maintains a high accuracy across all subgroups.

\begin{table}[ht!]
\centering
\footnotesize
\resizebox{\linewidth}{!}{%
\begin{tabular}{lcccccccccc}
\toprule
\bf Skin Tone & {\bf1} {\scriptsize[2]} & {\bf2} {\scriptsize[871]} & {\bf3} {\scriptsize[3008]} & {\bf4} {\scriptsize[522]} & {\bf5} {\scriptsize[184]} & {\bf6} {\scriptsize[85]} & {\bf7} {\scriptsize[54]}& {\bf8} {\scriptsize[49]}& {\bf9} {\scriptsize[6]}& {\bf10} {\scriptsize[1]}
 \\[3pt]
& 0.00\% & 0.11\% & 0.47\% & 1.53\% & 0.54\% & 1.18\% & 0.00\% & 0.00\% & 0.00\%  & 0.00\%  \\ \midrule
\bf Gender & \multicolumn{5}{c}{{\bf Predominantly Feminine} {\scriptsize[2437]}} & \multicolumn{5}{c}{{\bf Predominantly Masculine} {\scriptsize[3544]}} \\[3pt] 
  &       \multicolumn{5}{c}{0.53\%} & \multicolumn{5}{c}{0.85\%} \\
  \midrule
\bf Age Bucket & & \multicolumn{2}{c}{{\bf 0-2 yrs} {\scriptsize[17]}} & \multicolumn{2}{c}{{\bf 3-19 yrs} {\scriptsize[568]}} & \multicolumn{2}{c}{{\bf 20-59 yrs} {\scriptsize[4925]}} & \multicolumn{2}{c}{{\bf> 60 yrs} {\scriptsize[247]}} & \\[3pt]

 & & \multicolumn{2}{c}{0.00\%} & \multicolumn{2}{c}{0.00\%} & \multicolumn{2}{c}{0.77\%} & \multicolumn{2}{c}{0.81\%} & \\
\bottomrule
\end{tabular}}
\caption{Detection error rate for ``person'' in \NEWNAME using the subset of the MIAP dataset~\cite{miap_aies} that contain exactly a single individual in the image. \NEWNAME maintains a low error rate across all subgroups. Skin tone follows the Monk Skin Tone Scale~\cite{Monk_2019}. Numbers inside square brackets correspond to the size of each bucket.
}
\label{table:miap_rai}
\end{table}

\textbf{Limitations.}\quad The analysis carried out in this section is necessarily limited, since fairness is a societal concept that cannot be reduced to  statistical metrics. 
We expect RAI evaluations to evolve over time as new issues are detected and reported in the literature and additional datasets become available. Statistical analysis is only a single step and does not substitute for studying the broad and delayed impact of deployed models.

In addition, we rely in some parts on automated tools for inferring attributes, which are not perfectly accurate and can lead to a broad categorization of people that misidentifies real identities. We do not support the creation or application of classifiers for sensitive attributes, such as gender or ethnicity, based on visual indicators and  encourage readers to delve into the comprehensive work outlining their potential risks, such as~\cite{hamidi2018gender,keyes2018misgendering}, for further insight. Also, while we use perceived gender presentation in our analysis that is provided by the data (i.e. in CelebA and FairFace), we acknowledge that people may express their gendered identities in numerous other ways. 

In our evaluation, toxicity is predicted based on the generated captions only. However, without knowing the context of the image, this can introduce false positives.

\section{Conclusions} 

In this work we draw more insights from further scaling vision and language models.
We show that the scaling and the improved training recipe results in a model that substantially outperforms previous state-of-the-art models, leads to emergent behaviors and identifies further margins for improvements.
In particular, we report that the model achieves significant improvements at document, chart, and infographic understanding, captioning, visual question answering, counting, and performs well on few-shot (in-context) captioning, video captioning and question-answering, and object detection.

\section*{Acknowledgements} 
We would like to thank 
Sarah Laszlo, Kathy Meier-Hellstern, Caroline Pantofaru,
Susanna Ricco, Candice Schumann, Ken Burke, Simon Wang, Rachel Hornung, Yichang Chen, Utsav Prabhu,
Abhijit Ogale, 
Kristina Toutanova, Weicheng Kuo, Jihyung Kil, Xiangning Chen, Liang Chen,
Rich Lee, Elizabeth Adkison, James Cockerille, Eric Ni, 
Erica Moreira, Victor Gomes,
Jeremiah Harmsen, Claire Cui, Slav Petrov, Tania Bedrax-Weiss, Joelle Barral, Tom Duerig, Paul Natsev, Fernando Pereira, Jeff Dean, and Zoubin Ghahramani
for helpful discussions, feedback, and support.

\newpage
\appendix
\section{Additional Model Details and Examples}
\label{appendix:model_info_and_example}

\subsection{\NEWNAME Architecture Illustration}
\begin{figure}[ht!]
\centering
\includegraphics[angle=270,width=0.7\linewidth]{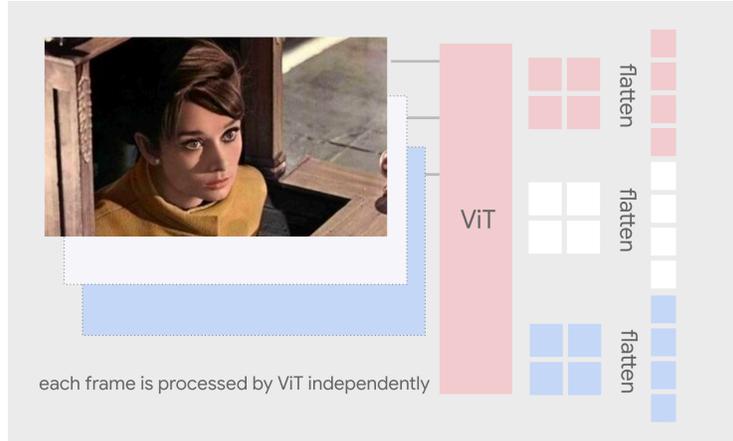}
\caption{Visual input for videos: each frame is independently processed by ViT; patch embeddings are flattened and concatenated together to form the visual representation.  (The example input image is in the \href{https://commons.wikimedia.org/wiki/File:Charadehepburn.jpg}{public domain}).}
\label{figure:video-input}
\end{figure}
\subsection{Tuning ViT-22B for better OCR capabilities}
The vision encoder's ability to understand text is crucial to several downstream tasks and general usability. JFT-based pre-training is insufficient to cover this, and so we tuned ViT-22B on WebLI-OCR data.
In order to stay true to the original discriminative classification-based objective used for ViT-22B, we turn OCR into a bag-of-words prediction task. OCR texts are tokenized using the mT5 tokenizer~\cite{xue-2021-mt5} across all languages, and the model is trained to predict whether or not a given token occurs in an image. This is treated as multilabel classification, with an expanded classification head.

In the ablation study shown in Table~\ref{table:ablation_vit22b_ocr}, we confirm that this this extra tuning step indeed has a significant improvement on Scene-Text understanding capabilities, demonstrated by the performance on ST-VQA and TextVQA.
Meanwhile, the performance on regular VQA tasks such as those in the VQAv2 benchmark also improves.

\subsection{Illustrative \NEWNAME Examples}
Table~\ref{appendix_fig:examples} shows representative examples of \NEWNAME, illustrating improved abilities related to counting (both of the simple and complex variety), in context text-reading capabilities, and spatial awareness.

\newcommand\rowincludegraphics[2][]{\raisebox{-0.45\height}{\includegraphics[#1]{#2}}}
\newcommand\imagecredit[2]{\small \textit{Image Credit: #1 [#2]}}
\newcommand\publicdomain[1]{\small \textit{\href{#1}{Public Domain Image}}}
\makesavenoteenv{tabular}
\makesavenoteenv{table}

\begin{table*}[ht!]
\resizebox{\linewidth}{!}{
\begin{tabular}{ll}
\toprule
\makecell[lt]{
\rowincludegraphics[width=0.5\linewidth]{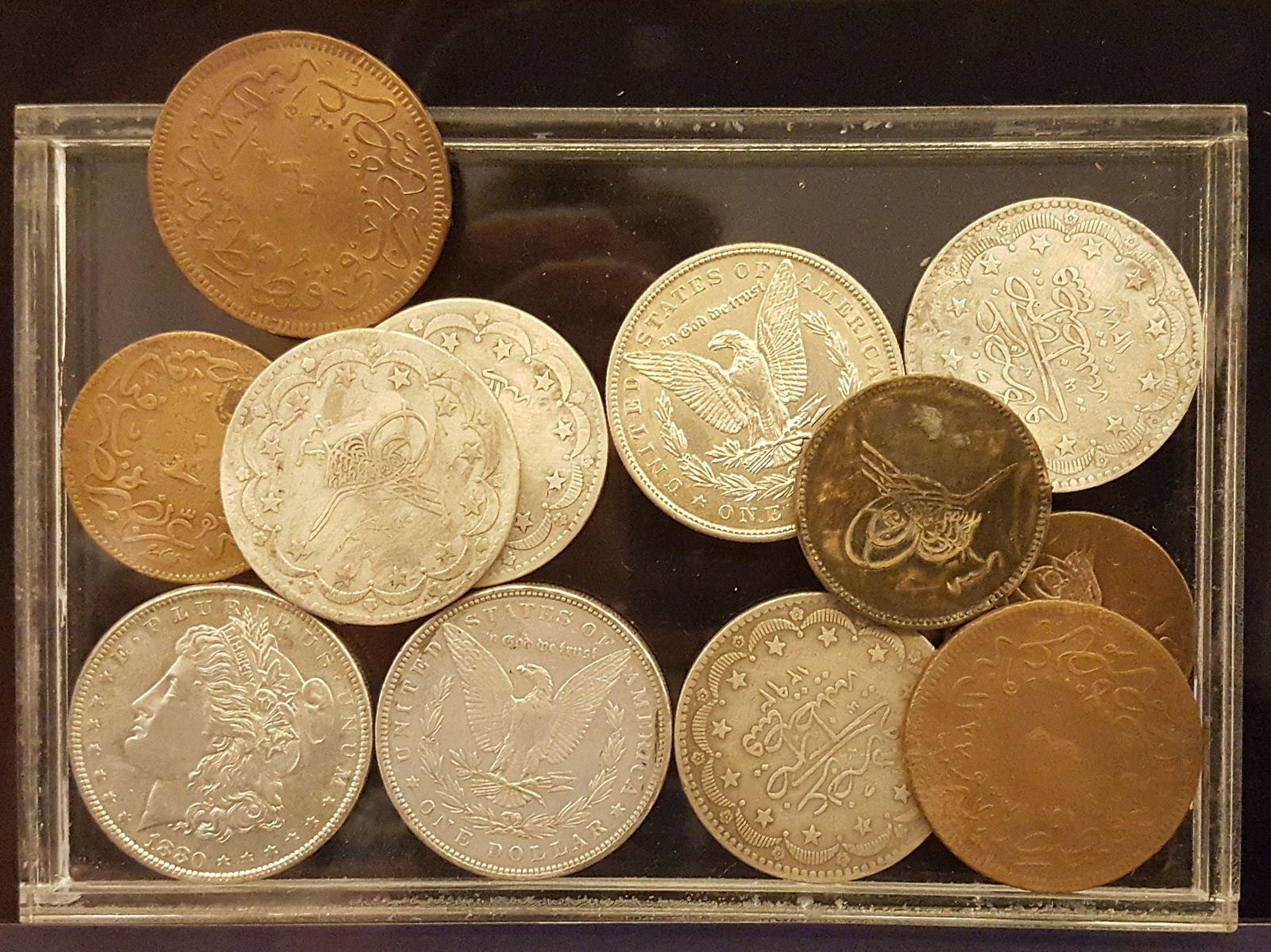} 
\\
\imagecredit{
\href{https://commons.wikimedia.org/wiki/File:Coins_of_19th_century.jpg}{Wikimedia Commons}}
{\href{https://creativecommons.org/licenses/by-sa/4.0/deed.en}{CC BY-SA 4.0}}
}
&
\makecell[lc]{
Q: how many coins are there? \\
A: 12 \\
\\
Q: how many one-dollar coins are there? \\
A: 2 \\
}
\\
\midrule
\makecell[lt]{
\rowincludegraphics[width=0.5\linewidth]{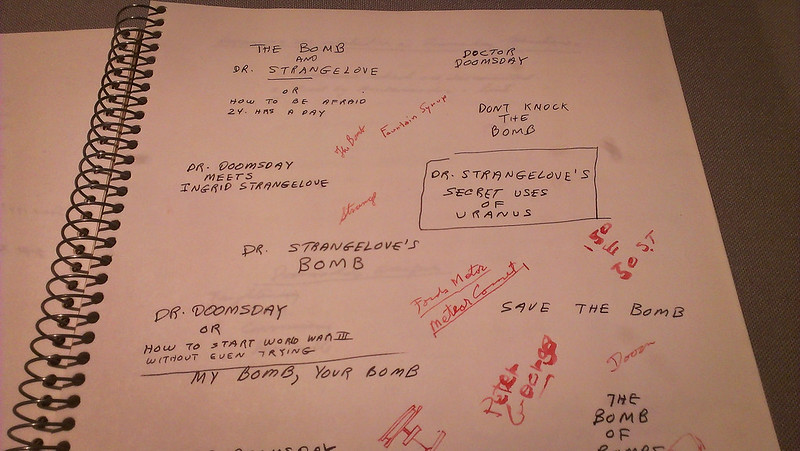} 
\\
\imagecredit{
\href{https://www.flickr.com/photos/chrisgold/}{ChrisGoldNY} (\href{https://www.flickr.com/photos/chrisgold/9169951242}{flickr})}
{CC BY-NC 2.0}}
&
\makecell[lc]{
Q: what is written inside the box? \\
A: dr. strangelove's secret uses of uranus \\
\\
Q: what is written on the top-left corner of the page? \\
A: the bomb and dr. strangelove \\
\\
Q: what is written on the top-right corner of the page? \\
A: doctor doomsday
}
\\
\midrule
\makecell[lt]{
\rowincludegraphics[width=0.4\linewidth]{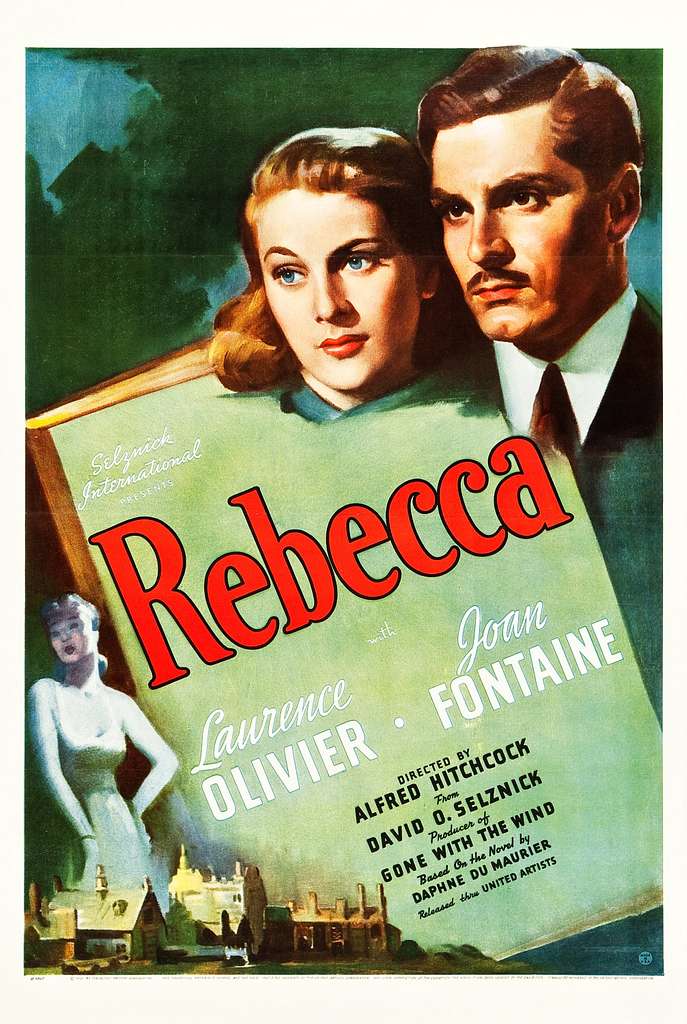} 
\\
\publicdomain{https://picryl.com/media/rebecca-1939-poster-b32552}
}
&
\makecell[lc]{
Captioning: \\
a movie poster for rebecca starring laurence olivier and joan fontaine. \\
\\
Q: who directed this movie? \\
A: alfred hitchcock \\
\\
Q: who produced this movie? \\
A: david o. seznick \\
}
\\
\bottomrule
\end{tabular}}
\caption{Examples of counting, text reading capabilities with context and spatial awareness.  Results are generated by the multi-task-finetuned models using the model's inherent OCR capabilities (i.e., without the use of an external OCR system).}
\label{appendix_fig:examples}
\end{table*}

\newpage
\section{Additional results: Image Captioning and VQA}
\label{appendix:image_results}
\subsection{Information of Downstream Image Benchmarks}
\begin{table*}[t!]
\centering
\resizebox{\linewidth}{!}{%
\begin{tabular}{lcll}
\toprule
Benchmark & Visual Domain & Description & Metric\\
\midrule 
COCO Captions & \multirow{11}{*}{Natural Images} & Captioning of natural images & CIDEr \\
NoCaps & & Captioning of natural images & CIDEr \\
TextCaps & & Captioning of natural images containing text & CIDEr \\
VizWiz-Cap & & Captioning of photos taken by people who are blind & CIDEr \\
VQAv2 && VQA on natural images & VQA accu. \\
OKVQA && VQA on natural images requiring outside knowledge & VQA accu. \\
TextVQA && VQA on natural images containing text & VQA accu.\\
VizWiz-QA && VQA on photos taken by people who are blind & VQA accu.\\
ST-VQA && VQA on natural images containing text & ANLS \\
TallyQA && VQA with counting questions & EM \\
OVEN && VQA on natural images for visual entity recognition & EM \\
InfoSeek && VQA on natural images for visual info-seeking questions & Relaxed EM\\
\midrule
OCR-VQA & \multirow{3}{*}{Illustrations} & VQA on images of book covers & EM\\
ChartQA && VQA on images of charts & RA\\
AI2D && VQA on images of scientific diagrams & EM \\
\midrule
DocVQA & \multirow{2}{*}{Documents} & VQA on images of scanned documents & ANLS \\
InfographicsVQA && VQA on images of infographics & ANLS \\
\midrule
Screen2Words & \multirow{2}{*}{UIs} & Captioning a UI screen to describe functionality & CIDEr \\
Widget Captioning && Captioning a UI component on a screen & CIDEr\\
\bottomrule
\end{tabular}}
\caption{Summary of Image Captioning and VQA benchmarks used for evaluating \NEWNAME}
\label{table:benchmark_summary}
\end{table*}
Table~\ref{table:benchmark_summary} summarizes the Image Captioning and VQA benchmarks. For benchmarks modeled only end-to-end without OCR pipeline input (Table~\ref{table:captioning} and Table~\ref{table:image_extended}), fine-tuning is performed with resolution \imres{672}. For Scene-Text and Document Understanding tasks presented in Table~\ref{table:scene-text-like}, fine-tuning is performed with resolution \imres{756}.
\subsection{Extended Tables of Image Benchmarks}
An extended table of results on some Image Benchmarks is shown as Table~\ref{table:image_extended}.
\begin{table*}[ht!]
\centering
\resizebox{\linewidth}{!}{%
\begin{tabular}{lcccccccccccc}
\toprule
& COCO && \multicolumn{2}{c}{NoCaps} && \multicolumn{2}{c}{VQAv2} && OKVQA && \multicolumn{2}{c}{TallyQA} \\
\cmidrule{2-2} \cmidrule{4-5} \cmidrule{7-8} \cmidrule{10-10} \cmidrule{12-13}
Model & Karp.-test & & val & test & & test-dev & test-std && val && simple & complex \\
\midrule
SimVLM & 143.3 && 112.2 & 110.3 && 80.03 & 80.34 && - && - & -\\
CoCa (2.1B) & 143.6 & & 122.4 & 120.6 && 82.3 & 82.3 && - && - & -\\
GIT (0.7B) & 144.8 & & 125.5 & 123.4 & & 78.56 & 78.81 && - && - & -\\
GIT2 (5.1B) & 145.0 & & 126.9 & \textbf{124.8} && 81.74 & 81.92 && - && - & -\\
OFA (0.9B) & 145.3 & & - & - && 82.0 & 82.0 && - && - & -\\
Flamingo (80B) & 138.1 & & - & - && 82.0 & 82.1 && 57.8$^*$ && - & -\\
BEiT-3 (1.9B) & 147.6 && - & - && 84.2 & 84.0 && - && - & -\\
PaLM-E (562B) & 138.7 && - & - && 80.0 & - && \textbf{66.1} & - & - \\
MoVie & - && - & - && 69.26 & - && - && 74.9 & 56.8 \\
\NAME(17B) & 149.1 & & \textbf{127.0} & 124.4 & & 84.3 & 84.3 && 64.5 && 81.7 & 70.9 \\
\midrule
\NEWNAME(55B) & \textbf{149.2} && 126.3 & 124.3 && \textbf{86.0} & \textbf{86.1} && \textbf{66.1} && \textbf{86.0} & \textbf{75.6}\\
\bottomrule
\end{tabular}}
\caption{Results on COCO Captions (Karpathy split), NoCaps, VQAv2, OKVQA, and TallyQA with end-to-end modeling without OCR pipeline input. The ``simple'' and ``complex'' are test subsplits.}
\label{table:image_extended}
\end{table*}

\subsection{Multi-lingual Captioning}
\paragraph{Multilingual captioning on XM-3600} The Crossmodal-3600 (XM3600) benchmark contains a geo-diverse set of 3600 images with human-annotated reference captions in 36 languages \cite{Thapliyal2022Crossmodal3600AM}.  Table \ref{table:i18n_captioning} presents multilingual results for both \NAME (current SoTA on XM-3600) and \NEWNAME, both finetuned with \imres{224} resolution.
Overall, \NEWNAME improves on the SoTA performance across 5 of the 7 languages we report here (and for 14 of the total 35 languages considered); notably, the performance on English is 4 CIDEr points lower compared to PaLI.
The 35-language average CIDEr score is in the same ballpark between \NAME and \NEWNAME, with a slight +0.5 advantage for \NAME. 
\begin{table*}[ht!]
\centering
\resizebox{0.77\linewidth}{!}{%
\begin{tabular}{lccccccccc}
\toprule
Model && en & fr & hi & iw & ro & th & zh & 35-lang avg. \\
\midrule
\NAME && \textbf{98.1} & 75.5 & 31.3 & 46.8 & 35.8 & 72.1 & \textbf{36.5} & \textbf{53.6} \\
\NEWNAME && 94.2 & \textbf{78.7} & \textbf{32.0} & \textbf{46.9} & \textbf{36.9} & \textbf{75.3} & 36.1 & 53.1\\
\bottomrule
\end{tabular}}
\caption{CIDEr scores on image captioning for the Crossmodal-3600 benchmark for seven diverse languages (English, French, Hindi, Hebrew, Romanian, Thai, and Chinese), as well as the average of the 35 languages covered by the benchmark. Both models are finetuned with \imres{224} resolution.} 
\label{table:i18n_captioning}
\end{table*}

\subsection{TallyQA and the emergence of complex counting capability}
\begin{figure}[ht!]
\centering
\includegraphics[width=0.6\linewidth]{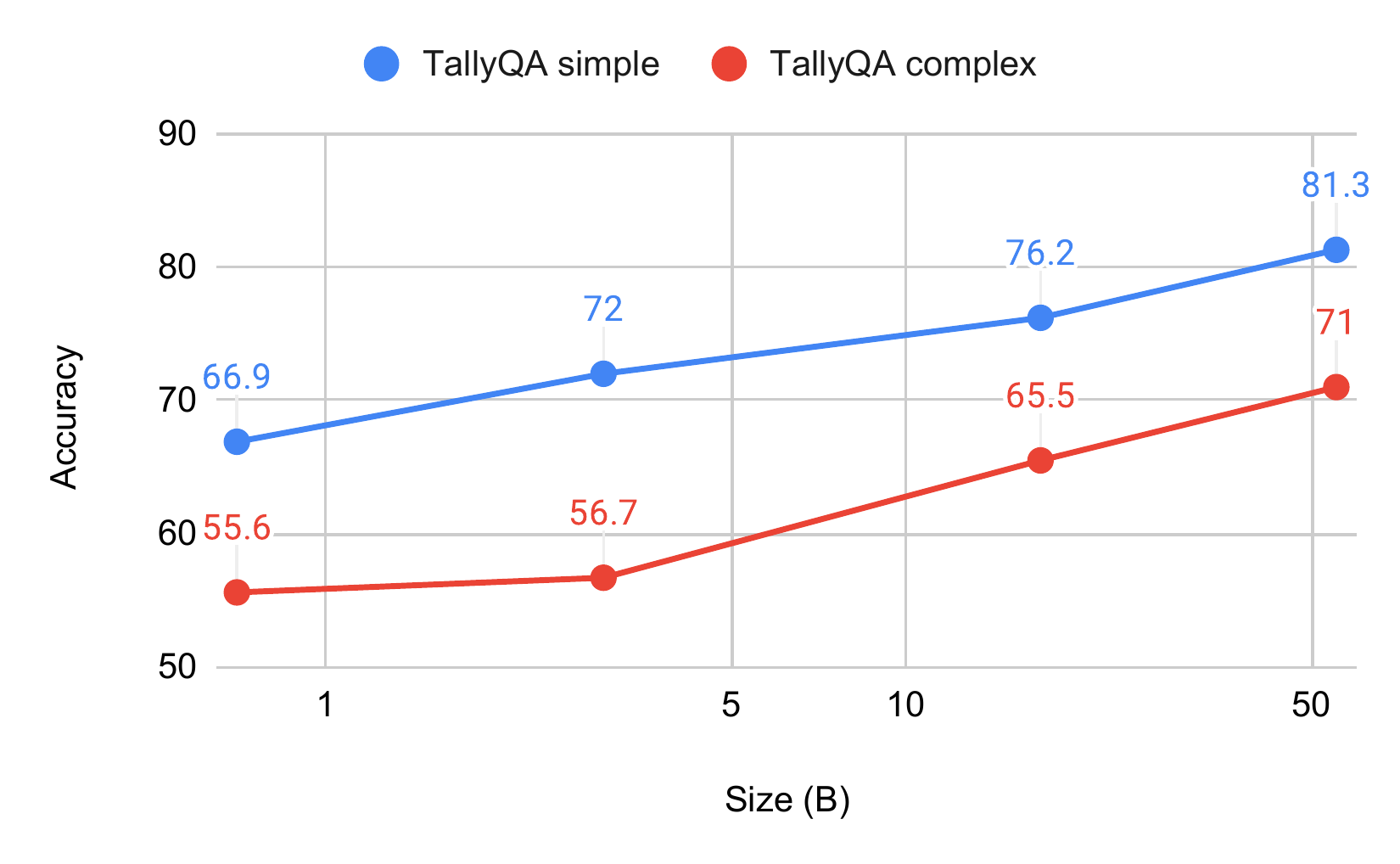}
\caption{Performance on TallyQA splits for simple and complex using \NAME variants and \NEWNAME. All models use \imres{224} image resolution. The emergent behavior on complex counting beyond the 3B size is made clear with \NEWNAME.}
\label{fig:tallyqa_vs_size}
\end{figure}

We present in Table~\ref{table:tallyqa_vs_size} the performance of similar models across a wide range of capacity -- from 700M parameters to 55B parameters for \NEWNAME.
The graphs in Fig.~\ref{fig:tallyqa_vs_size} illustrate how simple counting appears to follow a more linear progression as parameter-size increases, while complex counting appears to show emergence somewhere before the datapoint provided by the performance of PaLI 17B.
This corresponds to our intution that complex counting is a true multimodal task that requires additional capabilities from a model, in terms of the alignment that is required between the visual information and the prompt specification.

\begin{table*}[ht!]
\centering
\resizebox{0.9\linewidth}{!}{%
\begin{tabular}{lccc}
\toprule
Model & TallyQA simple & TallyQA complex & Weighted average\\
\midrule 
\NAME (700M) & 66.9 & 55.6 & 62.4 \\
\NAME (3B)  & 72.0 & 56.7 & 65.9 \\
\NAME (17B) & 76.2 & 65.5 & 71.9 \\
\NEWNAME (55B) & 81.3 & 71.0 & 77.2\\
\bottomrule
\end{tabular}}
\caption{Performance on TallyQA splits for simple and complex questions. All models use \imres{224} image resolution.}
\label{table:tallyqa_vs_size}
\end{table*}

\subsection{Details on Few-shot Modeling}
\subsubsection{Few-shot Formulation}
\begin{figure}[ht!]
\includegraphics[scale=0.32]{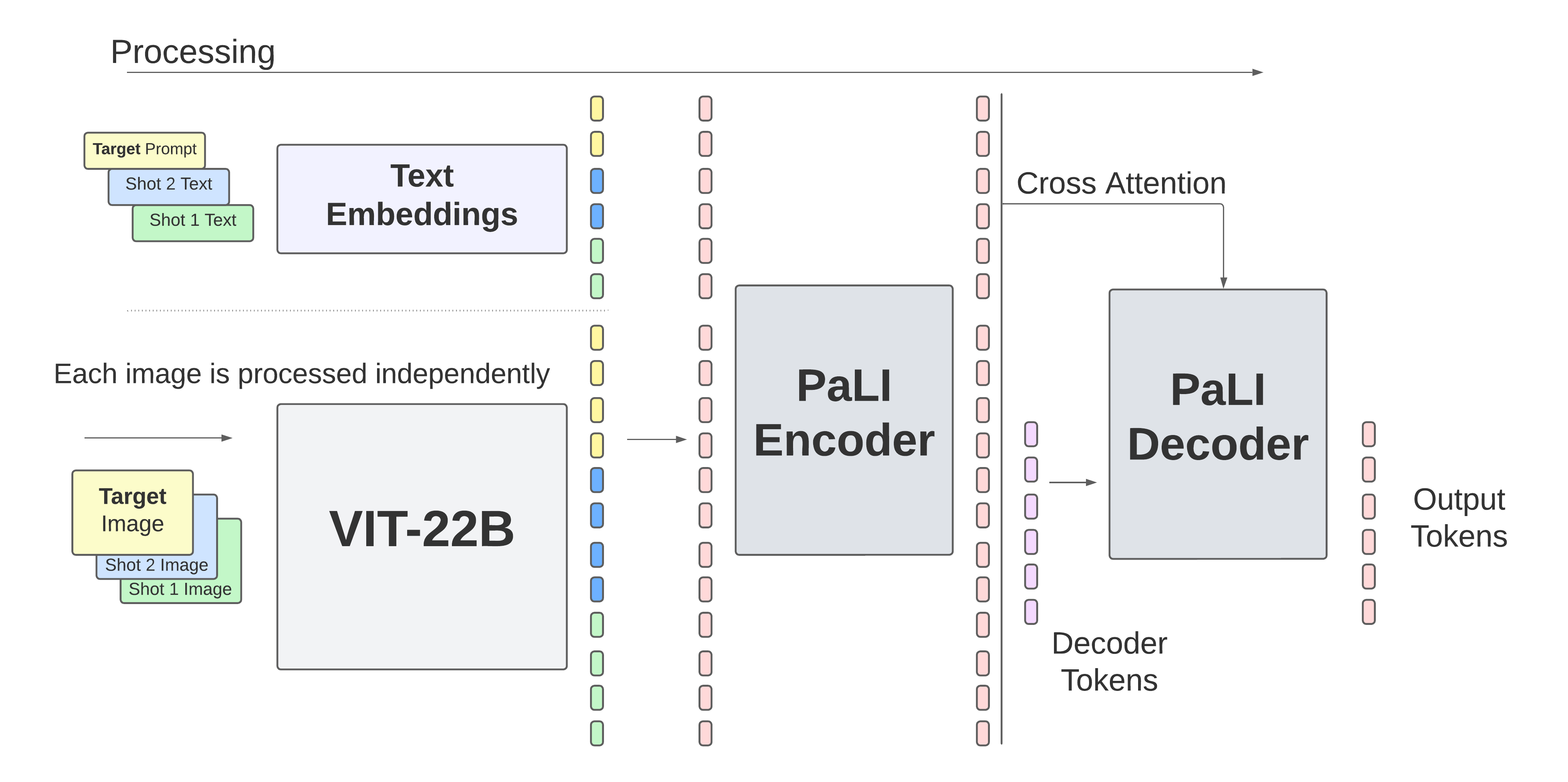}
\centering
\caption{A detailed view on how the few-shot exemplars are fed to the model components.}
\label{fig:fewshot_diagram}
\end{figure}

Figure \ref{fig:fewshot_diagram} illustrates the network flow of a few shot model. The text and prompt part of each shot is embedded and concatenated as text features for the \NEWNAME model.
Each shot's images and the target image are independently encoded by the ViT component, and the ViT features are concatenated along the sequence axis as visual features. Conditioned on that sequence, the \NEWNAME decoder autoregressively makes the predictions for the target image.

\label{sec:appendix-few-shot}

\paragraph{Encoder shot and Decoder shots} While images for all few-shot examples and target example are given as input to the model,
text information can be provided in different ways.  During inference time, all text information related to the few-shot examples is given to the encoder; in the case of a Multi-answer VQA task, for example, this includes both the prompts that contain the questions, and the expected answers.  Prompt for the target example is also given to the encoder, and the decoder is tasked with generating an answer for the target example.  During training, however, we increase the training efficiency by making the model predict answers for both the target example and selected shots (the {\em decoder shots}).  That is, we partition the $N$ shots in two sets: encoder shots ($N_e > 0$) and decoder shots ($N_d \ge 0$), such that $N_e + N_d \le N$. We use up to 4 shots in total during pre-training (i.e. $N = 4$ ), and sample $N_e$ uniformly at random from 1 to $N$.
Text input for encoder shots contain both prompts and answers.
The decoder shots, however, act as if they were target examples:
 their text input to the encoder contains only the prompt, and the decoder needs to predict answers for the decoder shots in addition to the target example.

\begin{wrapfigure}{r}{0.5\textwidth}
\includegraphics[scale=0.4]{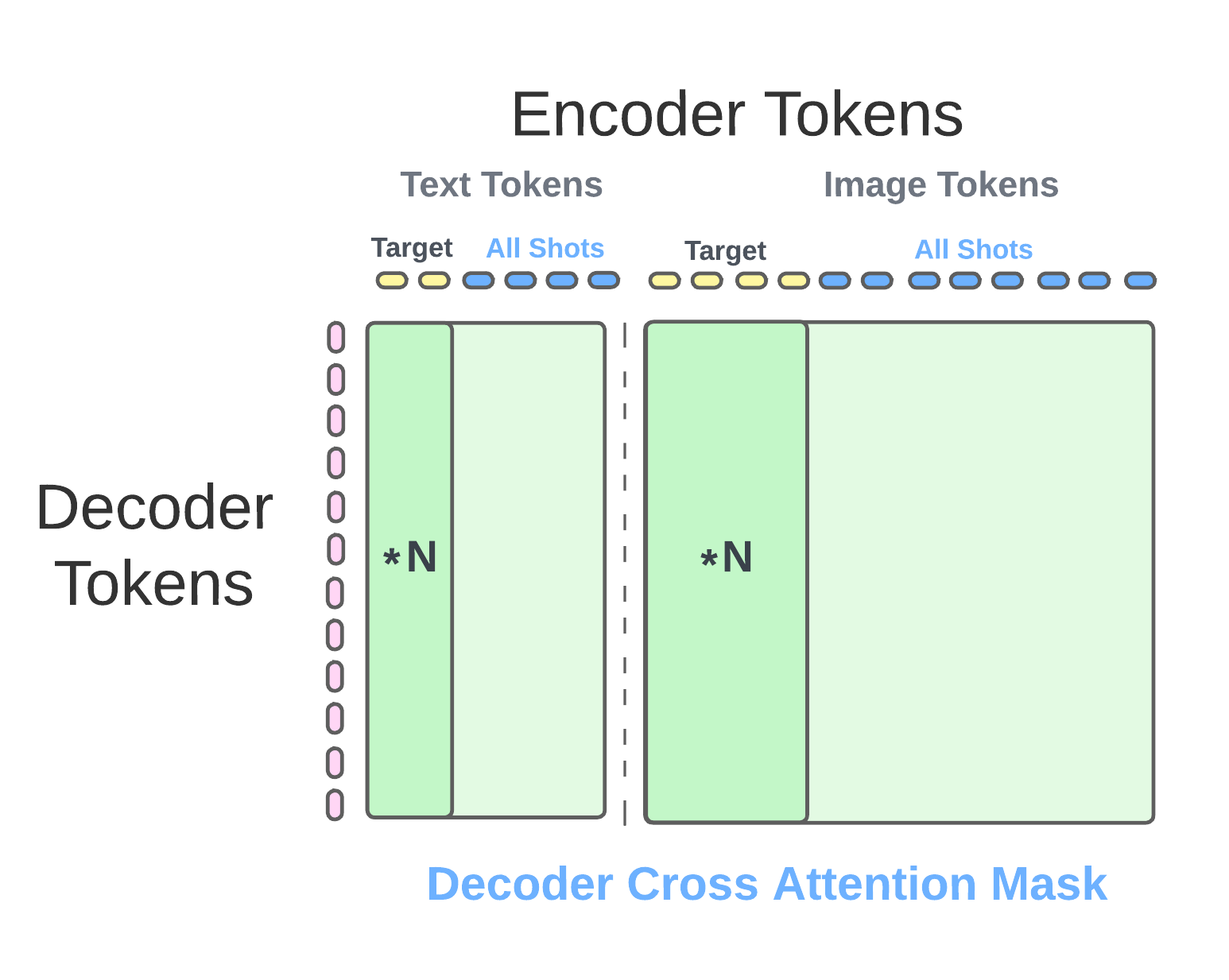}
\centering
\caption{Re-weighted attention with few-shots.}
\label{fig:fewshot_attention}
\end{wrapfigure}

\paragraph{Attention re-weighting} 
Increasing the number of shots turned out to be challenging,
potentially due to cross-attention to target example input tokens getting diluted by the large number of shots.
To address this, we introduce an attention re-weighting mechanism.
As shown in Figure~\ref{fig:fewshot_attention},
we explicitly boost the weights for cross attention between decoder tokens and encoded tokens from the target example (that is, the target image and the target text prompt).

Specifically, if there are $N$ shots in total, when decoding each token we multiply the cross attention weights by $N$ for the target image and text tokens from the encoder outputs.
We observe this attention re-weighting technique is especially helpful when we provide the model with many shots (e.g. 32 shots).
\cite{hao2022structured} introduces a technique along similar lines to manipulate attention weights when gathering them from different threads of encoded shots at inference time.

\begin{figure}[!ht]
    \centering
    \includegraphics[width=\linewidth]{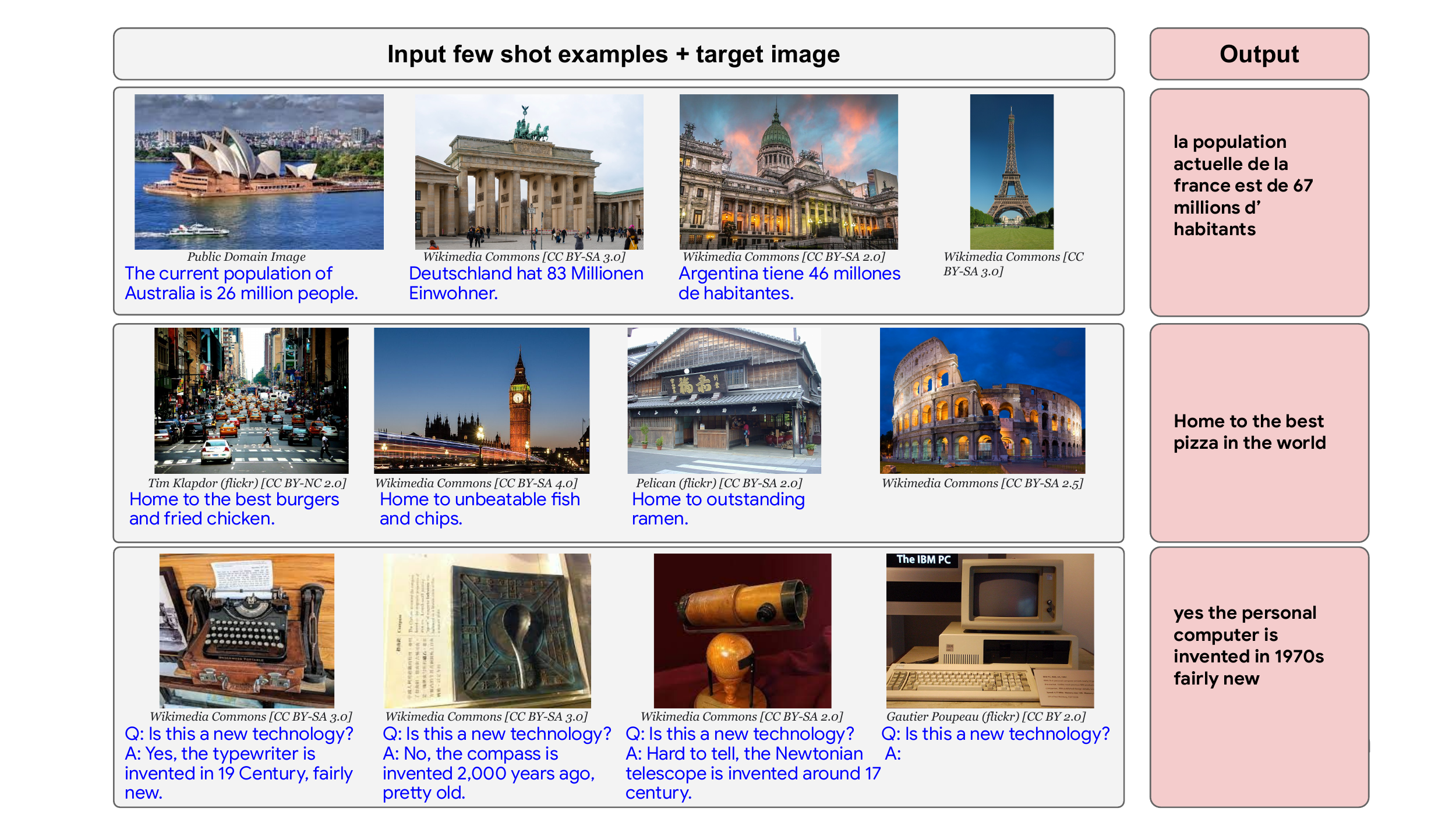}
    \caption{Qualitative Results on few-shot captioning (first two rows) and VQA (the last row) tasks.}
    \label{fig:qualitative-result-few-shot}
\end{figure}

\subsubsection{Additional Few-shot Results} 

\paragraph{Multilingual captioning results} 
Table~\ref{table:fewshot_xm3600_eval} reports the CIDEr scores for 7 languages and an average over 35 languages to demonstrate PaLI's multilingual captioning capabilities on the \XM benchmark in teh few-shot setting. 
The pre-trained model (no few-shot finetuning) achieves an average score of 22.7.
The \NEWNAME model achieves an average score of 45.1 for 4 shots and 47.1 for 32 shots.
Note that the 32-shot \NEWNAME average CIDEr score is only 6 points behind the fully finetuned model, which uses roughly 600k training examples per language (while the few-shot approach does not update the model parameters).

\begin{table}[ht!]
\centering
\resizebox{\linewidth}{!}{%
\begin{tabular}{l|cccccccc} \toprule
                  & \multicolumn{8}{c}{Crossmodal-3600 Captioning}   \\ \hline
   & en   & fr   & hi   & iw   & ro   & th   & zh  & 35-lang avg. \\\hline
 \NEWNAME 0-shot & 48.8 & 25.0 & 10.5 & 20.1 & 13.0 & 33.3 & 18.4 & 22.7 \\
 \NEWNAME (2 text-only shots\footnote{Equivalent with the Flamingo ``0-shot'' setting.}) & 54.5 & 46.7 & 12.0 & 22.2 & 9.4 & 40.3 & 23.7 & 25.8 \\
 \NEWNAME 4 shots & 77.8 & 62.5 & 22.2 & 38.7 & 30.2 & 56.0 & 27.7  & 45.1\\
 \NEWNAME 32 shots & 81.4 & 66.1 & 25.6  & 40.6 & 32.4 & 59.4 & 29.7 & 47.1 \\\hline
 \NEWNAME (finetuned) & 94.2 & 78.7 & 32.0 & 46.9 & 36.9 & 75.3 & 36.1 & 53.1\\ 
 \bottomrule
 \end{tabular}}
  \caption{Few-shot performance of the \NEWNAME model on multilingual captioning tasks.}
 \label{table:fewshot_xm3600_eval}
 \end{table}

\paragraph{Qualitative results} Figure \ref{fig:qualitative-result-few-shot} shows 3 examples on few-shot captioning and VQA tasks for qualitative analysis. The first row shows captions for the images using the images' original language, demonstrating the cross multilingual transfer of the few-shot capability.
The second row captions the images with a country's popular food, showing that the few-shot approach can access the model's world knowledge.
The last row shows a VQA with an explanation-like scenario where we ask if the technologies in the images are ``new''. Generally speaking, the shown personal computer was produced more than 40 years ago and could be regarded as old technology considering the fast pace of the current high-tech development. However, the 3 input shots provide the detailed calibration for the concept of ``new'' and the few-shot model successfully take the context and output ``new'' with plausible explanation to the very old PC.

\subsubsection{Few-shot ablation results}
In this section, we present and discuss some ablation results for few-shot we explored in order to inform our final design choices on \NEWNAME.
Unless otherwise specified, we use a 700M-parameter model with the same encoder-decoder architecture, consisting of a ViT-B/16 vision encoder and a mT5-base encoder-decoder language model. 

\paragraph{Pooling vs not pooling image tokens}
To mitigate the computational burden that arises with many shots, we can pool (for example, average) the per-image tokens before concatenating all input tokens. This pooled image tokens model achieved a CIDEr score of 56.3 for 4-shots COCO captioning, which is substantially lower than the full model's CIDEr score of 61.7.
This highlights the importance of keeping all the tokens coming out of the ViT encoder, despite the computational overhead.

\paragraph{Limited-range Encoding Attention.} 
We explore per-example image-text attention, as proposed and applied in \cite{alayrac2022flamingo}.
Under this approach, the image query tokens for each example can only attend to its corresponding text tokens, while the text query tokens can attend to all tokens. By using this per-example attention model, we achieved a CIDEr score of 59.6, which is 2.1 points lower than the full attention model's CIDEr score of 61.7 for 4-shots COCO captioning.

\paragraph{Attention re-weighting for large number of shots.} 
We report the few-shot results on COCO captioning from early-stopped PaLI-2 3B models; in this case, we did not apply normalized attention in training.
We provide the test results with and without attention re-weighting during \emph{inference} for a different number of encoder shots.
Attention re-weighting achieves increasing CIDEr scores of 82.1, 84.3 and 84.5 with 4, 8 and 16 shots respectively.
On the other hand, the model achieves 83.4, 76.5 and 66.3 without attention re-weighting. The decreasing performance may suggest that the model fails to locate the target image and text prompt among the large number of shots, whereas the attention re-weighting helps the model to focus on the target features.
Accordingly, we decided to include attention re-weighting during finetuning for \NEWNAME.

\paragraph{Distributing shots between encoder and decoder.}
We explore the use of both encoder and decoder shots during pre-training. We pretrain the PaLI-2 700M model on PaLI-2 mixtures with varying number of encoder shots (between 1 and 4).
The remaining shots (up to exactly 4) are used as decoder shots.
Using only encoder shots leads to a 64.0 CIDEr score for 4 shots in COCO captioning.
The best mix of encoder and decoder shots achieves a CIDEr score of 65.2. This suggests splitting shots leads to a more challenging pre-train task that helps the model learn more efficiently.

\subsection{Finetuning hyperparameters}
The hyperparamter choices for downstream finetuning experiments are summarized in Table~\ref{table:finetune_hyperparam}. As mentioned in the Main Text, for all of the downstream finetuning experiments, we used a reduced set of hyperparameters, without heavy per-task optimization.
\begin{table*}[ht!]
\centering
\resizebox{0.9\linewidth}{!}{%
\begin{tabular}{lccc}
\toprule
Benchmark & learning rate schedule & Steps before LR decay to 0 & batch size \\
\midrule
COCO & \multirow{6}{*}{linear decay from 1e-4} & 10k & 256 \\
VQAv2 && 20k & 256 \\
OCRVQA && 20k & 256 \\
Multitask-VQA && 20k & 256 \\
Multitask-Captioning && 20k & 256 \\
All other && 5k & 128 \\
\bottomrule
\end{tabular}}
\caption{Hyperparameter used for finetuning \NEWNAME.}
\label{table:finetune_hyperparam}
\end{table*}
\subsection{Multi-task finetuning}
We deduplicated every training set mixture over the test sets of every task in order to prevent leakage of any test-set examples into the training set. The mixture is formed by putting the training examples of each subtask together, with heuristic adjustments for a better balance. Following the resolutions for the single-task finetuning, the multi-task captioning and VQA finetuning are done with 672 and 756 image resolutions, respectively.
The multitask finetuning covers just about 5M examples, which is 20k steps with a batch size of 256.
For scene-text and document understanding tasks, the multi-task finetuning uses the end-to-end setting without OCR pipeline input.

The following aspects made multitask finetuning particularly challenging: (i) all tasks used the same prompt without task-specific indicators; the model is thus required to adapt to the style of multiple benchmarks simultaneously. 2) We do not perform per-task validation set optimization.
All subtasks are evaluated using the same checkpoint, but tasks converge to their optimal value at a different pace.
\begin{table*}[ht!]
\centering

\resizebox{\linewidth}{!}{%
\begin{tabular}{lc@{\hspace{0.15cm}}c@{\hspace{0.15cm}}c@{\hspace{0.15cm}}c@{\hspace{0.15cm}}c@{\hspace{0.15cm}}c@{\hspace{0.15cm}}c@{\hspace{0.15cm}}c@{\hspace{0.15cm}}c@{\hspace{0.15cm}}c@{\hspace{0.15cm}}}
\toprule
& VQA & OK & Text & VizWiz & ST & OCR & Info & Doc & Chart & \multirow{2}{*}{Avg.} \\
Model & v2 & VQA & VQA & VQA & VQA & VQA & VQA & VQA & QA & \\
\midrule
Split & test-dev & val & val & test-dev & val & test & test & test & test & - \\
\midrule
Previous Multi-task SOTA & 84.3 & 64.5 & 68.4 & 71.6 & 75.1 & 71.3 & 40.0 & 76.6 & 70.5 & - \\
\midrule
Single-task FT & \textbf{86.0} & \textbf{66.1} & \textbf{71.9} & \textbf{72.6} & \textbf{80.2} & \textbf{75.9} & 49.2 & 80.0 & \textbf{70.9} & -\\
Multi-task FT & 84.3 & 63.5 & 71.4 & 71.4 & 79.0 & 73.4 & \textbf{50.7} & \textbf{80.9} & 70.6 & -\\
Multi-task (+/-) & \red{-1.7} & \red{-2.6} & \red{-0.5} & \red{-1.2} & \red{-1.2} & \red{-2.4} & \textcolor{teal}{+1.5} & \textcolor{teal}{+0.9} & \red{-0.3} & \red{-0.8}\\
\bottomrule
\end{tabular}}
\caption{Scores from multi-task finetuning compared with those from single-task finetuning for VQA. Validation or test-dev set numbers are reported for some tasks.}
\label{table:multitask_vqa}
\end{table*}
\subsection{Ablation studies}
We first show in Table~\ref{table:ablation_vit22b_ocr} the advantage brought by the OCR co-training stage of ViT-22B. We pair the vanilla ViT-22B and the ViT-22B with additional OCR  co-training with a small language model mT5-base and pretrain these models on 40M of WebLI-OCR data with the splitOCR objective, before finetuning on ST-VQA. Co-training on image and OCR classification has a significant advantage on ST-VQA and TextVQA. In the meantime, the performance on VQAv2, which is not very scene-text heavy, is improved as well. Moreover, we found that making the top left patch white, which helped the co-training of image classification and ocr classification on ViT-22B, is not required for the subsequent training of \NEWNAME. 

For ablation of the \NEWNAME training procedure, we used a 5B model with UL2-3B and ViT-G with 2B parameters, which is roughly a 10:1 down-scale of the \NEWNAME 55B model. 
 
\begin{table*}[ht!]
\centering
\resizebox{\linewidth}{!}{%
\begin{tabular}{lccccc} 
 \toprule
 Model & OCR-task Indicator & ST-VQA & TextVQA & VQAv2 & 3-task avg. \\
 \midrule
 mT5-base + Vanilla ViT-22B & No & 42.6 & 36.1 & 68.9 & 49.2 \\
 \midrule
 mT5-base + ViT-22B-OCR & No & \textbf{47.0} & 38.9 & 69.8 & \textbf{51.9} \\
 mT5-base + ViT-22B-OCR & Yes & 46.2 & \textbf{39.4} & \textbf{70.2} &\textbf{51.9}\\
 \bottomrule
 \end{tabular}}
\caption{Advantage of the OCR co-training stage of ViT-22B. Pretraining is performed with resolution \imres{224} and finetuning is with \imres{448}. Numbers reported are on validation split.}
\label{table:ablation_vit22b_ocr}
\end{table*}

For stage 1 training, we show in Table~\ref{table:ablation-vqgan} that adding image token generation does not harm the performance on the main image+language understanding tasks. 
\begin{table*}[h!]
\centering
\resizebox{0.5\linewidth}{!}{%
\begin{tabular}{lcc} 
 \toprule
 Mixture & COCO & VQAv2 \\
 \midrule
 without ViT-VQGAN & 139.3 & 77.3 \\
 with 10\% ViT-VQGAN & 139.7 & 77.1 \\

 \bottomrule
 \end{tabular}}
\caption{Ablation experiment showing adding ViT-VQGAN tokens does not harm understanding performance (captioning and VQA tasks).}
\label{table:ablation-vqgan}
\end{table*}

\newpage
\section{Additional results: Video Captioning and QA}
\label{appendix:video_results}
Below we give a brief description of each video data set we used for evaluation. Note that we freshly collected the data when performing the experiments, which led to different effective numbers of videos in different splits in some cases, see Table~\ref{table:video-data-wipeout}.

These descriptions refer to the original dataset size, but we train on (sometimes significantly) fewer videos --- the exact numbers are given in Table~\ref{table:video-data-wipeout}.
This is because not all videos in the datasets were available online at the time of writing (e.g., due to user deletion).

\subsection{Datasets \& Benchmarks}
\textbf{MSR-VTT~\cite{xu2016msr}:} This dataset consists of 10K open domain video clips for video captioning, with 20 captions each.
The duration of each video clip is between 10 and 30 seconds. We follow the standard splits proposed by~\cite{xu2016msr} and report results on the test set. 

\textbf{VATEX~\cite{wang2019vatex}:} VATEX includes captions for 41K videos sampled from the Kinetics-600 dataset, with 10 English captions each.
We report results on the English public test set.

\textbf{ActivityNet Captions~\cite{krishna2017dense}:} This dataset consists of 100K temporally localized
sentences for 20k videos. 
We follow the standard split containing 50/25/25\% of the dataset for training, validation and testing, and use ground truth temporal proposals at evaluation following~\cite{krishna2017dense}.
Note that following other works~\cite{wang2021end}, we use the \texttt{val\_1} split for validation and \texttt{val\_2} split for testing.

\textbf{Spoken Moments in Time (SMIT)~\cite{monfort2021spoken}:} This dataset consists of long captions obtained via audio recordings for 500k short video clips. While this dataset has  been traditionally only used for text to video retrieval, we find that it is a strong benchmark for captioning as it is the largest manually annotated set of videos with text captions. 

\textbf{ActivityNet-QA~\cite{yu2019activitynet}:} The dataset contains 58,000 question-answer pairs for videos in the ActivityNet dataset~\cite{caba2015activitynet}. We report accuracy (using exact string match) on the test split. Note that we do open-ended generation for all VideoQA datasets.

\textbf{MSR-VTT-QA~\cite{xu2017video}:} This dataset was created using a semi-automatic pipeline on top of the MSR-VTT dataset. We report accuracy (using exact string match) on the test split.

\textbf{NExT-QA~\cite{xiao2021next}:} We focus on the Open-Ended QA task, which consists of 52,044 question-answer pairs for a total of 5,440 videos (sampled from the VidOr dataset\cite{shang2019annotating}). Exactly following Next-QA~\cite{xiao2021next} and Flamingo~\cite{alayrac2022flamingo}, we report the Wu-Palmer Similarity (WUPS) on the test set.

\begin{table}[h]
\centering
\footnotesize
\resizebox{\linewidth}{!}{%
\begin{tabular}{llrrrrrrr}
\toprule
{} & & MSR-VTT &  VATEX & ANet-Cap &    SMIT & M-V-QA & ANet-QA &  NExT-QA \\
\midrule
 &       train &    6513 &  25991 &                37421 &  481094 &     158581 &          32000 &   37523 \\
Original size      &  valid.&     497 &   3000 &                17505 &   14604 &      12278 &          18000 &    5343 \\
  &        test &    2990 &   6000 &                17031 &   3513 &      72821 &           8000 &    9178 \\
  \midrule
 &       train &    4768 &  22902 &                30982 &  481094 &     116943 &          28020 &   37523 \\
Dataset size  &  valid.&     327 &   2657 &                14604 &    8096 &       8215 &          15890 &    5343 \\
 &        test &    2144 &   5276 &                14234 &    3513 &      53014 &           7050 &    9178 \\
\midrule
 &       train &   73.21 &  88.12 &                82.79 &   100.00 &      73.74 &          87.56 &  100.00 \\
\% Remaining &  valid.&   65.79 &  88.57 &                83.43 &   100.00 &      66.91 &          88.28 &  100.00 \\
 &        test &   71.71 &  87.93 &                83.58 &   100.00 &      72.80 &          88.13 &  100.00 \\
\bottomrule
\end{tabular}}
\caption{We freshly collect the data sets from the respective data sources. In cases where there are multiple question-answer pairs per video we report the number of question-answer pairs. Similarly, for ActivityNet Captions we report the number of captions. Due to missing videos which were removed after the original data sets were defined, most of our data sets are missing 10\% of the videos or more.
}
\label{table:video-data-wipeout}
\end{table}

\newpage
\section{Additional results: Image Classification}
\label{appendix:img_classification_results}
\paragraph{Setup for zero-shot and finetuning evaluation}

The setup used for the experiments here uses the \NEWNAME model to generate directly the (English) class name using the captioning prompt.
The output is considered correct if it matches exactly the class name (apart from ImageNet-REAL, where we check if the class corresponding to the output is in the set of correct labels).

\paragraph{Zero-shot Evaluation results}
We use the same scoring technique as in \NAME~\cite{pali2} to evaluate \NEWNAME in zero-shot setting (without training on any Imagenet data).
We use the \NEWNAME model obtained after the first stage of training (using the base 224 image resolution).

The results are presented in Table~\ref{table:imagenet_zero_shot}. We compare the results to \NAME~\cite{pali2} - previous zero-shot generative SOTA, and Flamingo~\cite{alayrac2022flamingo} - another generative model of similar architecture with comparable  1-shot and 5-shot results. 
Overall, we report that the results between \NAME and \NEWNAME for 0-shot are similar.

\begin{table*}[ht!]
\centering
\resizebox{0.85\linewidth}{!}{%
\begin{tabular}{lccccccc}
\toprule
Model (ImageNet data) & INet & REAL & INet-R &  INet-A  & INet-Sketch &  INet-v2 & ObjNet\\
\midrule
Flamingo-80B (1-shot) & 71.9      & - &     -       &  -      &  -    &  -  & -  \\
Flamingo-80B (5-shot) & 77.3 &  -  & - & -    &  -    &  -    & -\\
\NAME (17B) (0-shot)  & \textbf{72.11} & \textbf{76.43}  & 81.97   &  44.70   &  \textbf{63.83}  &   \textbf{64.46}  & 42.62 \\
\midrule

\NEWNAME (0-shot)  & 71.16  & 75.75 &  \textbf{82.96} &  \textbf{46.13}  & 61.58  & 63.91
    & \textbf{44.58}  \\

\bottomrule
\end{tabular}}

\caption{Top 1 accuracy results of 0-shot image classification on ImageNet~\cite{deng2009imagenet}, \mbox{ImageNet-REAL}~\cite{imagenet_real}, \mbox{ImageNet-R}~\cite{hendrycks2021many}, \mbox{ImageNet-A}~\cite{hendrycks2021nae}, \mbox{ImageNet-Sketch}~\cite{wang2019learning}, \mbox{Imagenet-v2}~\cite{recht2019imagenet} and ObjectNet~\cite{objectnet}.} 
\label{table:imagenet_zero_shot}
\end{table*}

\paragraph{Finetuning}
To test image classification capabilities, we finetune \NEWNAME on ImageNet~\cite{deng2009imagenet} and evaluate the resulting model on \mbox{ImageNet-REAL}~\cite{imagenet_real} and out-of-distribution datasets: \mbox{ImageNet-R}~\cite{hendrycks2021many}, \mbox{ImageNet-A}~\cite{hendrycks2021nae}, \mbox{ImageNet-Sketch}~\cite{wang2019learning}, \mbox{ImageNet-v2}~\cite{recht2019imagenet}.

We use the model from the first training stage (at resolution 224) and the one from the last training stage (at resolution 756).
We use the same training hyperparameters for all of runs (selected without any hyperparameter tuning).

The results can be seen in Table~\ref{table:imagenet_training_appendix}. 
We compare the results to generative model with open vocab -- GiT2~\cite{wang2022git} (using 384 image resolution), which is the current SOTA for full-finetuning on ImageNet. \NEWNAME achieves close to SOTA results for generative models on Imagenet, and other datasets.

\begin{table*}[ht!]
\centering
\resizebox{0.85\linewidth}{!}{%
\begin{tabular}{lccccccc}
\toprule
Model (resolution) & INet & REAL & INet-R &  INet-A  & INet-Sketch &  INet-v2 \\
\midrule
GIT2 (384)     & \textbf{89.22}  &   - &   -       &  -      & -    & -     \\
PaLI 3B   (224) & 85.11 & 88.71     &      \textbf{81.11}       &  45.71      &  70.00    &  78.23  \\
PaLI 17B (224) & 86.13  &   88.84 &   78.21       &  50.00      & 71.21    & 78.91     \\
\midrule
\NEWNAME (224)  & 88.22 & 90.36 & 77.66  &  55.97  &  72.56   &  81.42     \\
\NEWNAME (756) & 88.82 &  90.80   &  79.97    & \textbf{73.47}    &  \textbf{73.39}    & 83.48 \\
\NEWNAME$^\dagger$ (756) & 89.19 &  \textbf{90.98}   &  80.06   &  72.57    &  73.37    & \textbf{83.66} \\
\bottomrule
\end{tabular}}
\caption{Classification (top-1) accuracy with Imagenet~\cite{deng2009imagenet} fine-tuning on: ImageNet, \mbox{ImageNet-REAL}~\cite{imagenet_real}, \mbox{ImageNet-R}~\cite{hendrycks2021many}, \mbox{ImageNet-A}~\cite{hendrycks2021nae}, \mbox{ImageNet-Sketch}~\cite{wang2019learning}, \mbox{Imagenet-v2}~\cite{recht2019imagenet} (resolution in parentheses). \NEWNAME$^\dagger$ fine-tuned for 2.2x more steps.} 
\label{table:imagenet_training_appendix}
\end{table*}

\newpage
\section{Object Detection}
\subsection{Object detection as a VLM task}
\begin{figure}[ht]
\centering
\includegraphics[width=0.6\textwidth]{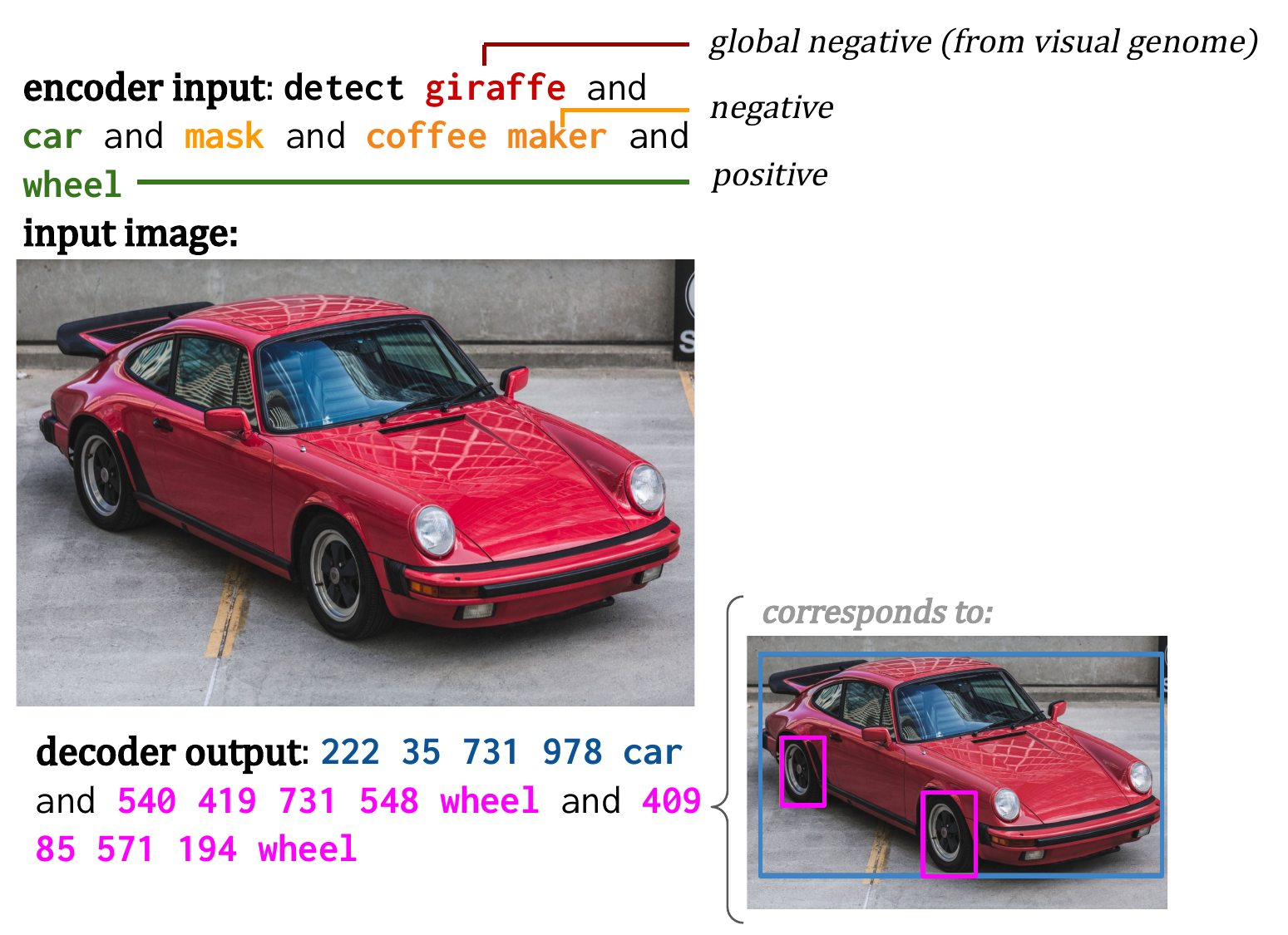}

\footnotesize{\textit{Image credits: Matthew Henry, burst, https://burst.shopify.com/photos/vintage-red-porsche}}

\caption{An example training pair, consisting of the text prompt, the image and the expected output. The prompt consists of multiple classes; we show a hypothetical Open Images V4 example, with positives `car' and `wheel', negative `giraffe` and global negatives `mask' and `coffee maker' (sampled from the visual genome label space).}
\label{app:fig:det_eg}
\end{figure}

Object detection is framed similarly to Pix2seq~\cite{chen2022pix2seq}, with two key differences: the use of a natural language vocabulary, and class-conditioning.
Prompt classes are fed to \NEWNAME's text encoder, in the format \texttt{detect class1 and class2 and class3}. The model is trained to only output bounding boxes corresponding to classes in this prompt.
We represent bounding boxes as coordinates in the same style as pix2seq~\cite{chen2022pix2seq}; that is, 4 integers \texttt{y\textsubscript{min}} \texttt{x\textsubscript{min}} \texttt{y\textsubscript{max}} \texttt{x\textsubscript{max}} ranging from 0 to 999. Figure~\ref{app:fig:det_eg} shows an example input.

\paragraph{Prompt sampling hyperparameters}
During training, a prompt for each example. We construct prompts from three pieces of information:
\begin{itemize}
    \item \textit{Positives}: These are the bounding boxes for objects definitely present in the image. During training, per example we sample $p^+ \sim \mathcal{U}(0, P^+_\mathrm{max})$, and keep that proportion of positives.
    \item \textit{Negatives}: These are the known instance negatives i.e. bounding boxes for objects definitely not present. For exhaustively labelled datasets like COCO, this is simply classes not labelled as positives. For non-exhaustively labelled datasets like LVIS, these are the classes not labelled as positives, which were presented to raters. During training sample $f^- \sim \mathcal{U}(0, 5.0)$, and use up to $f^-\times n^+$, where $n^+$ is the number of positives after sampling $p^+$.
    \item \textit{Global negatives}: These are negatives which are not explicitly labelled as negatives. They are taken from a wider label space combining multiple detection datasets. For a given example, valid global negatives consist of classes from the wider label space not explicitly labelled as positives or negatives. During training, we sample $f^{GN} \sim \mathcal{U}(0, 5.0)$ and append $f\times n^+$ global negatives, where $n_+$ is the number of positives after sampling $p^+$.
    
    By default, the combined label spaces of Visual Genome, Objects365 and OpenImagesV4 was used as the global label space, with the exception of detection finetuning, where LVIS and COCO label spaces were also added.
\end{itemize}
We truncate the number of total classes to $n_\textrm{max}$. $n_\textrm{max}$ and $P^+_\mathrm{max})$ are tuned per dataset to meet sequence lengths. Afer truncatation, we shuffle classes in the prompt.

\subsection{Preprocessing}
During pre-training, data is preprocessed to remove all LVIS-rare labels, following the protocol of OwlViT~\cite{owlvit}. This is not done for detection finetuning.
Images are randomly flipped horizontally, and randomly resized to between 0.3 and 2.0 $\times$ their original sized, followed by selecting a random square crop of the current training resolution. If the image is resized to be smaller than the current resolution, it is left as is. Images are finally padded to a square.

\subsection{Licenses and attribution for images used in Main Text Figure~\ref{fig:det_examples}}
\label{appendix:detection_information}
\begin{itemize}
    \item{Watermelon:} Credit: Sarah Pflug \\ \texttt{https://burst.shopify.com/photos/cutting-watermelon}.
    \item{Bowls: \\ \texttt{https://www.flickr.com/photos/ariesandrea/502826051/} CC-BY-NC-ND 2.0}
    \item{Business cat} Credit: Sarah Pflug, \\ \texttt{https://burst.shopify.com/photos/business-cat-in-office}
    \item{Wall} Credit: Matthew Henry \\ \scriptsize{\texttt{https://burst.shopify.com/photos/man-walking-in-front-of-this-is-paradise-wall?c=urban-life}}
\end{itemize}
\bibliographystyle{unsrt}
\bibliography{main}
\end{document}